\documentclass{article}
\PassOptionsToPackage{numbers, compress}{natbib}
\PassOptionsToPackage{table}{xcolor}
\usepackage[preprint]{neurips_2026}
\usepackage{xspace}
\usepackage{graphicx}
\usepackage{xcolor}
\usepackage{natbib}
\usepackage{caption}
\usepackage[title]{appendix}
\usepackage[most]{tcolorbox}
\usepackage{soul}
\usepackage{adjustbox}
\usepackage{booktabs}
\usepackage{amssymb}
\usepackage{float}
\usepackage{tcolorbox}
\usepackage{wrapfig}
\usepackage{enumitem}
\usepackage{hyperref}
\usepackage{cleveref}
\usepackage{multirow}
\usepackage{array}
\usepackage{arydshln}
\usepackage{tikz}
\usepackage{adjustbox}

\usepackage{placeins}
\usepackage[table]{xcolor}
\usepackage{booktabs}
\usepackage{csvsimple}
\usepackage{subcaption}

\def\equationautorefname~#1\null{Eq.~(#1)\null}

\newcommand{\bench}{\texttt{blind-spots-bench}\xspace}    
\newtcolorbox{promptbox}[1][]{
  colback=gray!10, colframe=gray, boxrule=0.5pt,
  arc=2mm, left=2mm, right=2mm, top=1mm, bottom=1mm,
  fontupper=\small\ttfamily, breakable,
  title=#1
}  

\newcommand{\toggle}[2][gray]{%
  \tikz[baseline=(X.base)]{
    \node[
      fill=#1!15,
      draw=#1!60,
      rounded corners=5pt,
      inner xsep=4pt,
      inner ysep=2pt,
      font=\small\ttfamily
    ] (X) {#2};
  }%
}

\title{Blind-Spots-Bench: Evaluating Blind Spots in Multimodal Models}

\author{%
  \textbf{Matteo Santelmo}\thanks{Equal contribution. Correspondence to \texttt{<matteo.santelmo@epfl.ch>},  \texttt{<xiuying.wei@epfl.ch>}. Matteo led the evaluation and Xiuying led dataset cleaning and labeling.  Dagger$^\dagger$ denotes advising roles.} \quad
  \textbf{Xiuying Wei}\footnotemark[1] \quad
  \textbf{Israa Fakih}\footnotemark[1] \quad
  \textbf{Felix Bauer}\footnotemark[1] \\
  \textbf{Juan Garcia Giraldo}\footnotemark[1] \quad
  \textbf{Chengkun Li}\footnotemark[1] \quad
  \textbf{Etienne Bamas}\footnotemark[2] \quad
  \textbf{Emmanuel Abbé}\footnotemark[2] \\[0.6em]
  École Polytechnique Fédérale de Lausanne (EPFL), Switzerland
}

\begin{document}

\maketitle

\vspace{-1.5em}
\begin{center}
\href{https://huggingface.co/datasets/matsant01/blind-spots-bench}{%
  \raisebox{-0.15em}{%
    \includegraphics[height=1.1em, trim=0 30 0 6, clip]{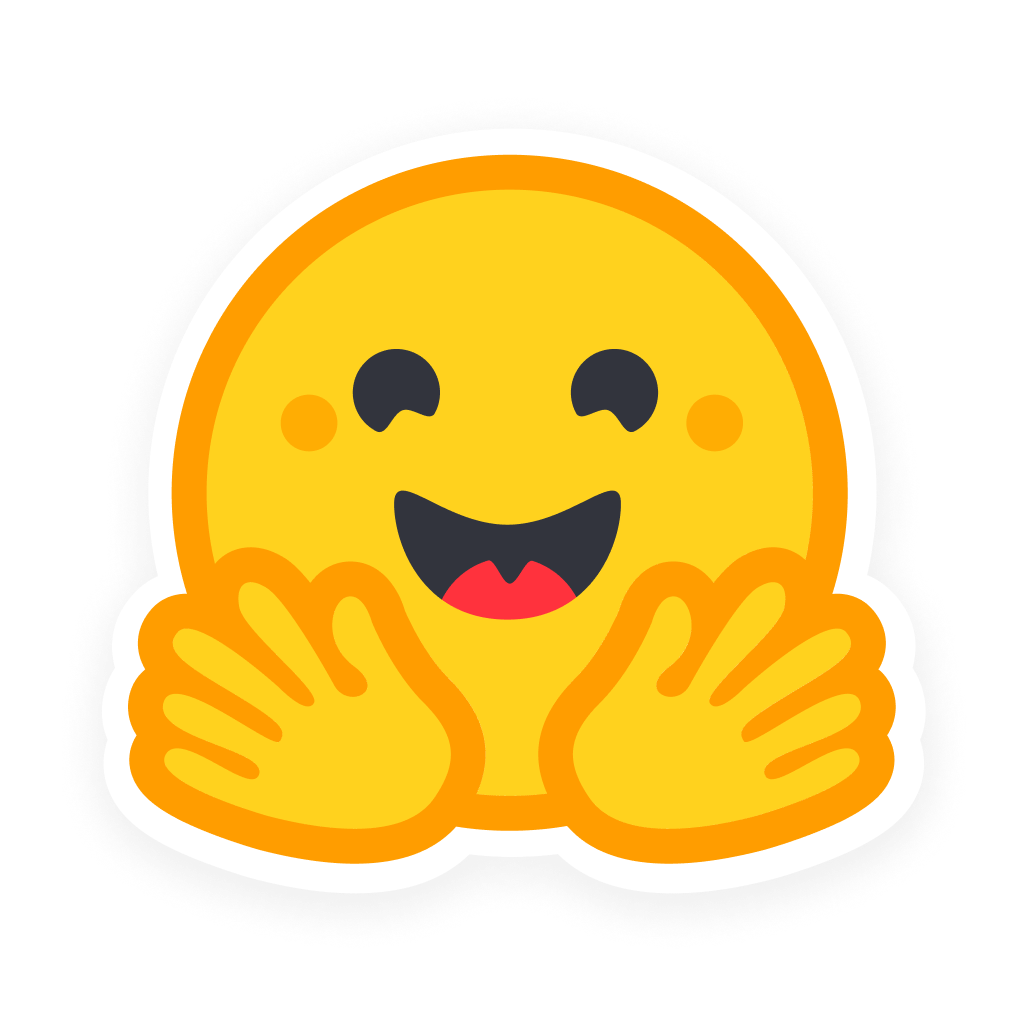}%
 }\,
  \texttt{blind-spots-bench}
}
\quad
\href{https://github.com/matteosantelmo/reasoning-blind-spots}{%
  \includegraphics[height=0.9em, trim=0 0 0 0, clip]{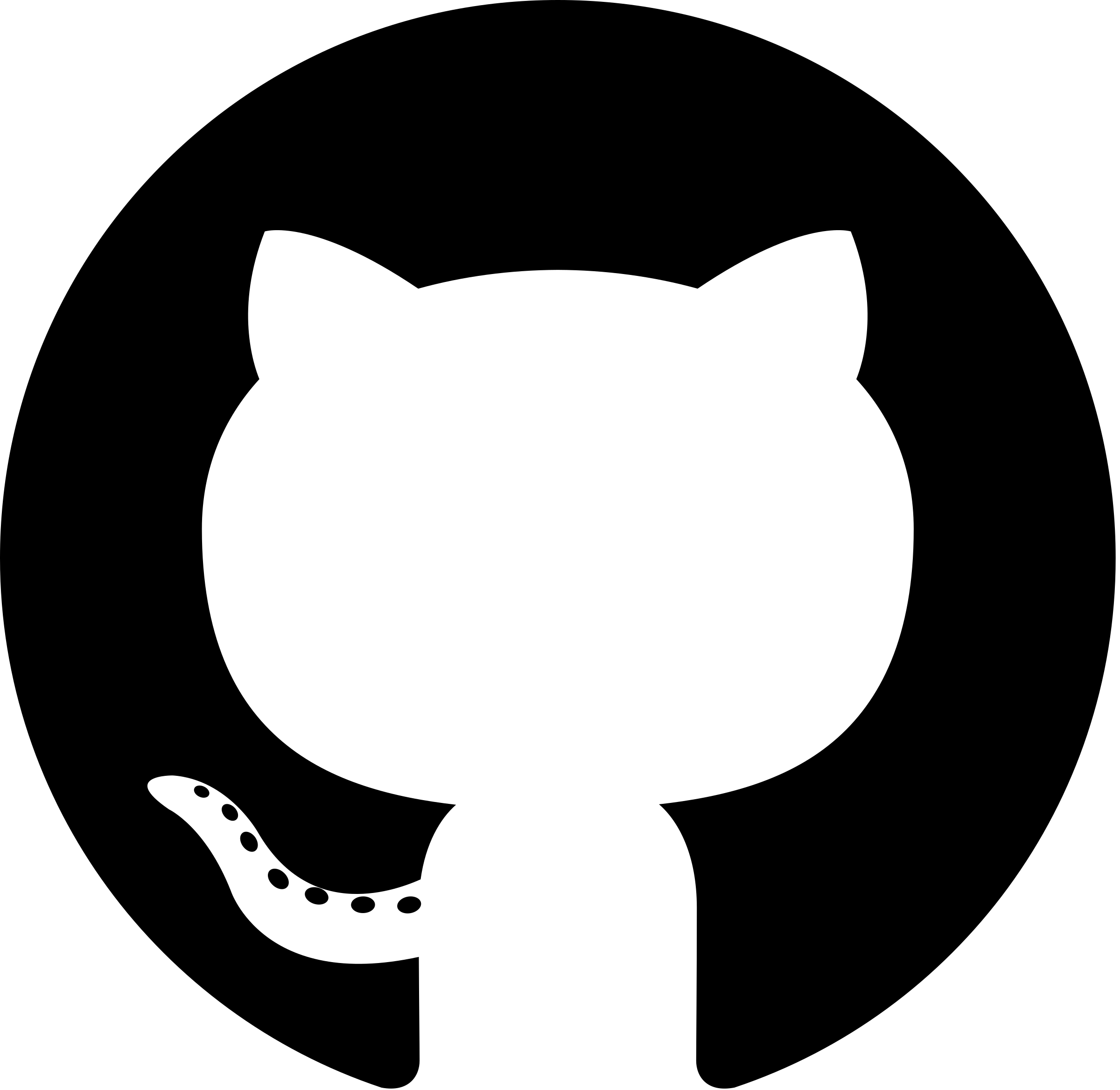}\,
  \texttt{eval-pipeline}
}
\end{center}

\begin{abstract}
Modern AI models achieve strong performance on many established benchmarks, yet they still fail on tasks that humans find almost trivial, such as manipulating a string or drawing a dog with five legs. These examples suggest that existing benchmarks may under-measure persistent blind spots in current systems. We introduce \bench, a benchmark designed to expose such blind spots through tasks that appear simple for humans but remain challenging for modern AI. We collect raw questions from students in an AI course, clean and annotate them with structured reference solutions, and propose a task taxonomy tailored to the resulting dataset of 235 samples. We further develop an automated grading pipeline to evaluate a wide range of models, including open-weight and closed-source language, vision-language, and image-generation models. Our analysis on \bench reveals that closed-source frontier models can substantially outperform open-weight models with even $\approx$10\% gap, even when they attain comparable performance on existing benchmarks. A more fine-grained analysis shows that no single model dominates across all task types, and that some tasks remain challenging for all evaluated models.
These results highlight the value of \bench as a diagnostic stress test for identifying concrete weaknesses in current modern models.
\end{abstract}

\begin{figure}[htbp]
    \centering
    \includegraphics[width=\linewidth]{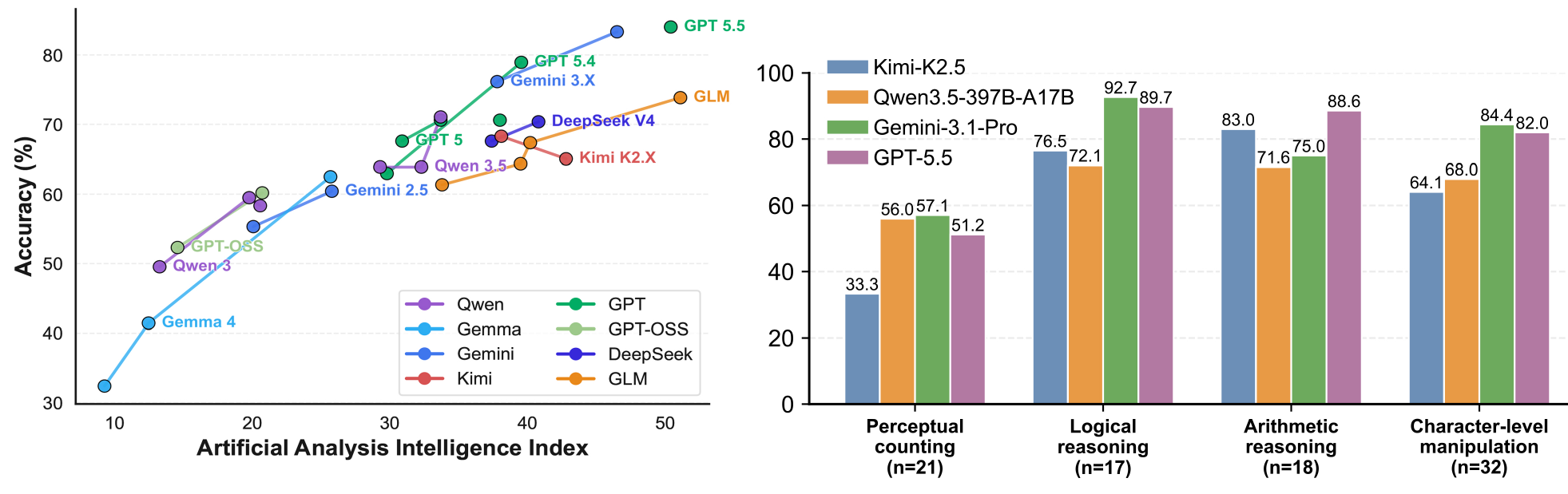}
    \caption{\textbf{Left}: Accuracy on \bench\ vs. Artificial Analysis Intelligence Index score for \textit{text-only} problems. \textbf{Right:} Performance of four VLM models on several sub-tasks.}
    \label{fig:acc_vs_aaii_text} 
 \end{figure}
\section{Introduction}


The rapid ascent of large language models (LLMs) and multimodal models has been marked by strong performance across a wide range of tasks, including mathematics, coding, reasoning, and vision. Recent frontier models~\cite{deepseek2026v4,openai2026gpt55,google2026gemini31} have achieved impressive results, often nearly saturating several carefully designed benchmarks~\cite{ifbench,gpqa_bench,tau_bench,swe_bench,matharena}, in some cases rivaling or surpassing human experts. Similar progress has also been observed on vision-language benchmarks, where modern systems demonstrate strong capabilities~\cite{mathvista,mmbench,mmmu_bench}.

In practice, however, these models still exhibit surprising failures on tasks that are straightforward for humans~\citep{zhou2023instruction,seely2025sudoku,tong2024eyes,fu2024large,chen2026babyvision}. For instance, a model may struggle to generate a string of a specified length~\cite{xiong2025enhancing}, produce an image of a cat with exactly four eyes~\cite{wiedemer2023compositional,binyamin2025make}, draw a clock showing a requested time~\cite{saxena2025lost}, or solve a relatively simple Sudoku puzzle~\cite{constraintbench}. These failures point to persistent blind spots across a range of abilities like spatial reasoning~\cite{spatialab}, logical consistency~\cite{liu2025evaluating}, and character-level manipulation~\cite{xiong2025enhancing}. 

To systematically study which questions remain easy for humans yet difficult for today's state-of-the-art models, we introduce \bench, a benchmark of 235 manually curated questions across multiple input-output format, accompanied by a reproducible evaluation framework. The questions are collected from students in a graduate-level AI course, who were asked in October 2025 to propose problems frontier AI chatbots failed to answer correctly. Each problem is annotated with a structured reference solution, question format, and task label. To support fine-grained analysis, we introduce a task taxonomy with three broad categories and finer-grained subtask labels: object-centric tasks, such as counting, recognition, and spatial reasoning; abstract reasoning tasks, mainly involving mathematical and logic-oriented problems; and language-and-knowledge tasks, involving linguistic and knowledge-based reasoning. We further develop an end-to-end evaluation pipeline to automatically grade answers against the reference solutions. 
Under this framework, we evaluate a broad range of frontier systems, including text-only LLMs, vision-language models (VLMs), and specialized image-generation models, covering 38 models in total.


Our results provide a systematic comparison across model variants and offer two complementary analyses. At the model level, the best-performing closed-source systems achieve approximately 10 per cent higher accuracy on \bench than the highest-scoring open-weight models. This difference remains marked even between models with comparable performance on established benchmarks (see \autoref{fig:acc_vs_aaii_text}). In contrast, open-weight models can be more effective under the same evaluation budget. We also find that tool use is not uniformly beneficial and can sometimes lead to lower accuracy. At the task level, our taxonomy supports detailed analyses of model capabilities. Fine-grained visual perception tasks, such as counting objects in images, remain challenging for all models with no more than 60\% accuracy results. No single model remains top-1 across all tasks. For example, the strongest \texttt{GPT} and \texttt{Gemini} models show different strengths across different tasks (e.g., 88.6\% of \texttt{GPT} on arithmetic reasoning). Meanwhile, other models, can still excel in specific categories, such as \texttt{DeepSeek-V4} on character-level manipulation and \texttt{Kimi-2.5} on arithmetic reasoning. Finally, we find that scaling model size within the same family does not always improve performance consistently.

We summarize our contributions as follows:
\begin{enumerate}[leftmargin=*]
    \item We introduce \bench, a curated dataset of 235 open-ended problems designed to expose blind spots in multimodal AI systems through tasks that are easy for humans but challenging for today's models. We apply a detailed cleaning and annotation process and provide structured reference solutions to support reliable evaluation.

    \item We develop an end-to-end evaluation pipeline, covering response generation, AI-based automatic grading, and manual auditing of grading reliability. Using this pipeline, we perform a comprehensive evaluation of 32 LLMs/VLMs, and 6 specialized image-generation models.

    \item We propose a task taxonomy and conduct fine-grained analyses of model behavior across task types. The taxonomy consists of three high-level categories and 12 subcategories, enabling us to examine question composition and sources of difficulty. Our analysis reveals both shared weaknesses, such as object counting, and model-specific strengths and failure patterns.
\end{enumerate}

\section{Related Works}

\paragraph{General and Multimodal Evaluation Benchmarks.}
A growing body of work has emphasized the importance of moving beyond a single aggregate benchmark score toward more structured characterizations of model reasoning errors. Recent benchmarks focusing on broad multidisciplinary reasoning performance, such as RBench~\cite{rbench_2025}, BIG-Bench~\cite{srivastava2023beyond}, GPQA~\cite{gpqa_bench}, and MMMU~\cite{mmmu_bench}, aggregate metrics across different categories, but fail to understand the differences in model behavior across specific reasoning components and identify residual weaknesses on simple but precise skills. In contrast, \bench introduces a structured framework that decomposes reasoning into fine-grained sub-tasks, allowing for a more interpretable assessment beyond aggregate scores, revealing that different models exhibit distinct strengths and weaknesses

\paragraph{Taxonomies and Failure Analysis.}
Moving beyond aggregate benchmarks, prior work has explored characterizing model behaviors through behavior testing, diagnostic challenge sets, and structured task taxonomies and failure analysis. In NLP, checklist~\cite{ribeiro2020beyond} introduced capability-level behavioral testing, while diagnostic benchmarks such as HANS~\cite{mccoy2019right} showed that models with high benchmark accuracy can still rely on brittle heuristics. Dynamic and contrastive evaluation frameworks further highlight the value of constructing targeted examples around model failures~\cite{kiela2021dynabench,gardner2020evaluating}. In multimodal evaluation, benchmarks such as MME~\cite{fu2023mme0}, ME-Reasoning~\cite{mme_reasoning_2025}, MM-Vet~\cite{yu2023mm}, GLUE~\cite{wang2018glue}, SIV-Bench~\cite{kong2025siv} organize evaluation around structured capabilities including perception, OCR, spatial reasoning, object-attribute understanding, social reasoning, multimodal logical reasoning, and distinctions between informal and formal reasoning or embodied settings~\cite{reasoning_failure_survey_2025}. Although these work provide useful high-level analyses of model behavior, they are either too coarse or too broad to support fine-grained analysis. As a result, they are difficult to apply to our benchmark setting for reasoning failure analysis. In this work, we instead propose a structured decomposition or reasoning problems into fine-grained sub-tasks and introduce failure modes for questions with more than one sub-task, where the type of failure is determined based on the model’s responses. This enables a more detailed and contextual analysis of errors.

\paragraph{Robustness Evaluation}
Another line of work studies whether high benchmark scores reflect robust reasoning. GSM-Plus \cite{li2024gsmplus} and Math-Perturb \cite{MATHPerturb2025} generate perturbed versions of problems from well established reasoning benchmarks. GSM-Symbolic \cite{Mirzadehetal2025} and VarBench \cite{qian2024varbench} focus on creating parametric templates to test performance invariance on equivalent problems.
These benchmarks complement \bench, but differ from it in their purpose. Perturbation benchmarks create controlled variants of existing problems, whereas \bench is built from naturally occurring failures of frontier models focusing on underrepresented tasks.

\section{Benchmark construction and Composition analysis}
\begin{figure*}[t!]
    \centering
    \includegraphics[width=\linewidth]{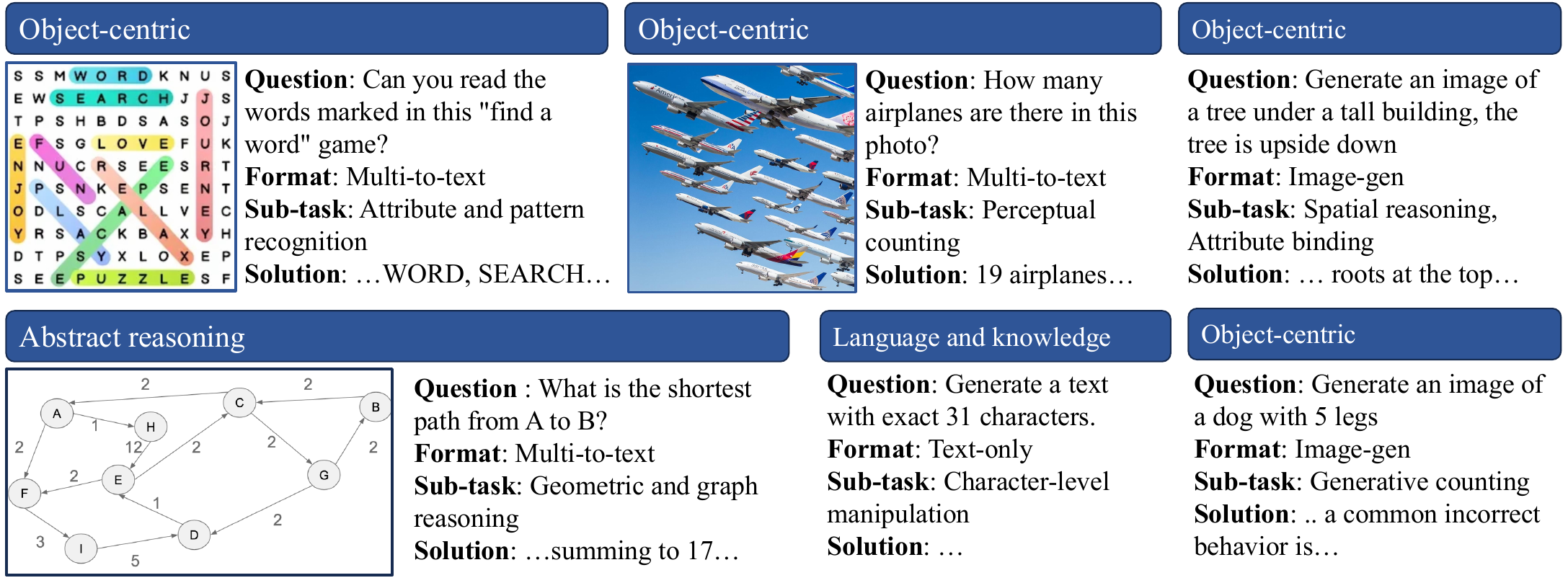}
     \caption{Representative examples from the dataset. While these tasks are generally easy for human, we find that they remain challenging for frontier models. Each example is annotated with its solution (an abbreviated version here), question format, task and sub-task categories, illustrating the diversity of skills evaluated in the benchmark. Full taxonomy and additional examples in \autoref{appendix:examples}.}
\label{fig:example}
\end{figure*}
\subsection{Data collection} \label{subsec:data_collection}
We collected raw questions from students in a graduate-level AI-related course.
Each student was asked to propose five questions that appeared easy to humans but were failed by frontier models available around October 2025.
This course-wide collection yielded approximately 287 raw questions. We then transformed the raw collection into a standardized benchmark through a systematic cleaning and annotation pipeline.
This process included filtering out overly difficult or duplicate entries, normalizing question formats, and ensuring each problem is clear and well-posed.
We further added key annotations as described below to facilitate verification and analysis. The resulting dataset, \bench, contains 235 samples and is publicly available on Hugging Face.\footnote{\url{https://huggingface.co/datasets/matsant01/blind-spots-bench}}
Representative examples are shown in \autoref{fig:example}.

\paragraph{Solution and Question format annotation}

For each question, we manually curate structured reference solutions to support reliable evaluation of model outputs. These solutions are designed to support downstream automatic verifiers, which require explicit and operational correctness criteria. Each reference solution specifies the expected answer, the necessary conditions for correctness, and common failure modes. The common failure modes are identified from two sources: errors observed in students' submitted interactions with AI agents during data collection, and additional error cases proposed by at least three annotators for each question. 

We also annotate each example with its question format to facilitate model comparison, since different models support different input-output formats. The benchmark includes three question-format categories: \textit{text-only}, where both input and output are text; \textit{image generation}, where the output is an image; and \textit{multi-to-text}, where the input contains both image and text and the output is text. These labels allow us to assign examples to appropriate models, compare systems within their applicable settings, and identify which formats current models struggle with most. 

\begin{figure}[htbp!]
    \centering

    \begin{minipage}[t]{0.23\linewidth}
        \centering
        \includegraphics[width=\linewidth]{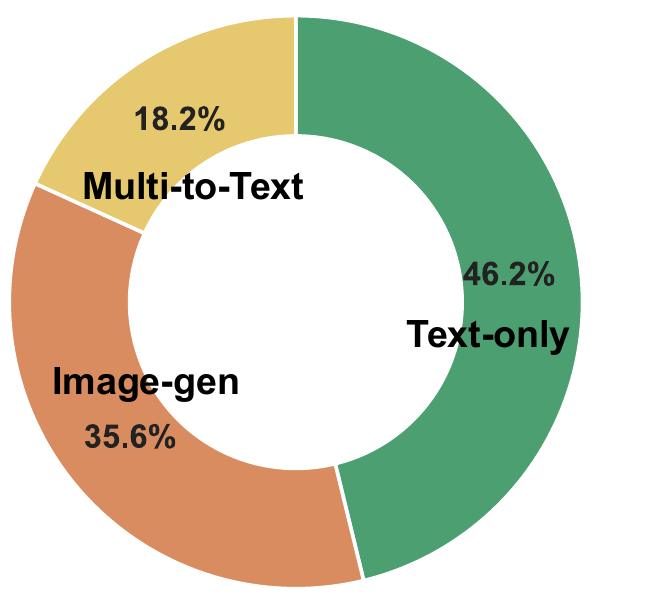}
    \end{minipage}
    \begin{minipage}[t]{0.75\linewidth}
        \centering
        \includegraphics[width=\linewidth]{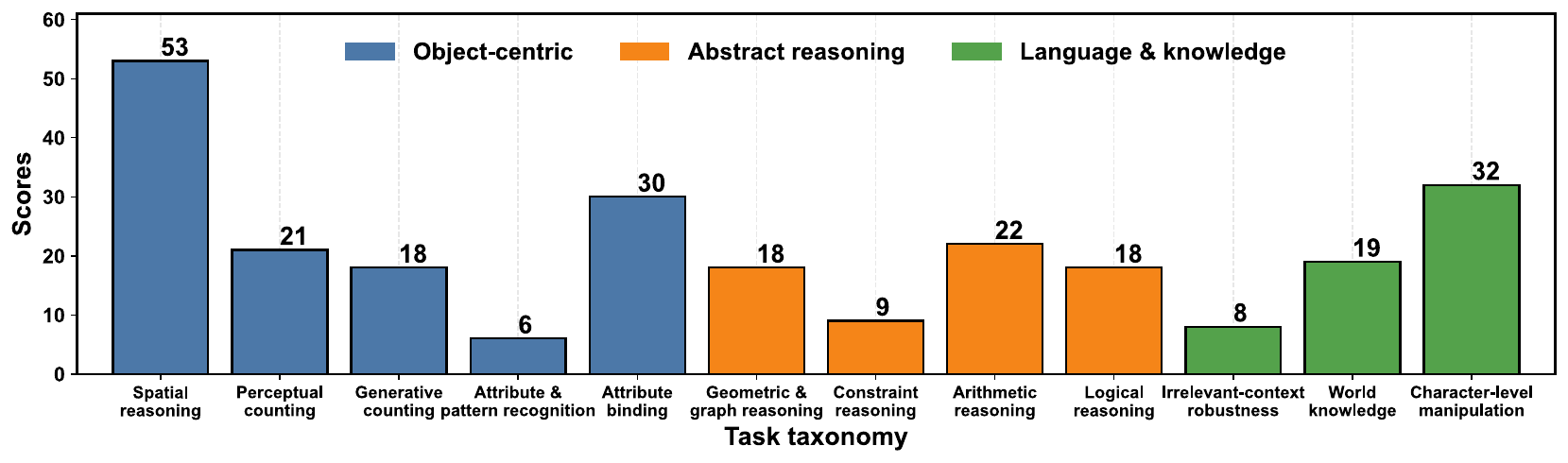}
    \end{minipage}
    \caption{\textbf{Left:} Question format composition.
\textbf{Right:} Task category composition.
Some questions~(about 15) involve multiple subtask categories; for these cases, we count one occurrence for each applicable subtask.
The bar chart reports the total number of occurrences for each fine-grained task category, grouped into three major categories. }
    \label{fig:dataset_composition}
\end{figure}

\paragraph{Task taxonomy}
Unlike many task-specific benchmarks~\cite{longbench,chartqa,spatialab}, \bench is designed to challenge AI systems and uncover reasoning blind spots across diverse question types, spanning a broad range of skills. To better understand the sources of difficulty and support subsequent analysis of model performance, we introduce a taxonomy tailored to our dataset.
Our taxonomy consists of three main categories: \emph{Object-centric}, \emph{Abstract reasoning}, and \emph{Language and knowledge}. Each category is further decomposed into sub-tasks targeting specific abilities. For example, as shown in \autoref{fig:example}, \emph{Object-centric} tasks cover abilities such as attribute and pattern recognition, spatial reasoning, perceptual counting, and generative counting.  We provide detailed descriptions of all categories and sub-tasks in \autoref{tab:dataset_category}, and discuss the overall distribution shown in ~\autoref{fig:dataset_composition} in the next section.


\paragraph{Review and Quality Control}
The dataset underwent a three-stage review process. First, all the questions were manually verified to confirm that each task is well defined and easy to interpret. Then we performed difficulty thresholding, removing questions that are easily solved by models or overly difficult for humans so that the benchmark focuses on blind spots. Finally, some questions were refined to clarify linguistic ambiguities and to ensure the correctness and completeness of the ground truth annotations.

\subsection{Dataset Composition}
Following the construction and annotation phase, we conduct a composition analysis of \bench dataset for both question format and task-category types. 
As illustrated in \autoref{fig:dataset_composition}, text-only questions constitute the largest portion of the dataset (46.2\%), followed by image-gen (35.6\%) then multi-to-text (18.2\%)  which are relatively fewer. 

Beyond the question format distribution, we further analyze the task and sub-task category composition in \autoref{fig:dataset_composition}. Since the prompts were designed to challenge frontier models, frequent sub-tasks can reveal sources of difficulty to some extent, rather than merely reflecting dataset statistics. They also align with known limitations of current models, suggesting that our taxonomy is both data-driven and grounded in prior findings. For example, spatial reasoning is the most frequent sub-task, appearing 53 times, consistent with prior observations that vision-language models struggle with spatial relations and visual transformations~\cite{stogiannidis2025mind,zhangvision}. Perceptual counting and generative counting are also common, reflecting the difficulty of counting visual instances~\cite{lvlm-count}, generating outputs with precise numerical constraints~\cite{binyamin2025make}, and binding objects to novel quantities~\cite{wiedemer2023compositional}. Logical reasoning often requires multi-step inference, where errors can accumulate when models make locally plausible decisions without sufficient global verification~\cite{saparov2022language}. Character-level manipulation is another challenge requiring exact length control or repeated token sequences, which may be partly explained by limitations of subword tokenization~\cite{zhang2024counting,fu2024large}. We provide a more detailed discussion in \autoref{app:task_type_analyses}.

While these individual sub-tasks have been studied separately in prior work, our contribution is to organize them into a unified taxonomy tailored to our dataset. This taxonomy connects dataset composition, known model weaknesses, and later model performance analysis.

\section{Experiments}
\subsection{Experimental Setup}
\paragraph{Evaluation Pipeline}

A harness was designed to automate the evaluation and scale it to a wide range of AI models on \bench. The evaluation pipeline consists of two steps: First, a \textit{solver} model is prompted to answer questions from \bench, without neither in-context examples nor chain-of-thought prompting. Subsequently, questions and responses are processed by a second model, that we refer to as \textit{grader}, to produce binary outputs. As the questions are designed to have objectively correct and incorrect answers, no partial grades are needed. The prompt for the \textit{grader} includes the original question, the solver's attempt, and the solution as described in \autoref{subsec:data_collection}. The full \textit{grader} prompts are reported in Appendix~\autoref{appendix:full_prompts}.

The pipeline is built using the Inspect AI \cite{UK_AI_Security_Institute_Inspect_AI_Framework_2024} framework (MIT License). As a \textit{grader} model we use \texttt{gemini-3-flash} which offers a good trade-off between grading quality, speed and cost. We enable the \textit{grader} with code execution to enhance its precision in verifying hard constraints, such as counting or finding sub-strings. The code of the evaluation pipeline is made public.\footnote{\url{https://github.com/matteosantelmo/reasoning-blind-spots}}

\paragraph{Grader Validation}

Since the evaluation relies on an AI \textit{grader}, we manually validate its agreement with human judgments.
Specifically, we sample more than 100 generated responses, equally drawn from different \textit{solver} models, score them using our pipeline, and in parallel manually annotate their correctness. This validation set is designed to include roughly equal numbers of responses graded as "correct" and "incorrect", preventing class imbalance from biasing agreement metrics. Full human-\textit{grader} agreement scores are reported in \autoref{tab:pipeline_validation}, with an accuracy of 96.6\% on tasks with textual outputs, and 90.9\% on the more challenging image-generation tasks.

Since our evaluation uses \texttt{gemini-3-flash} as the grader, we additionally assess it exhibits bias that inflates scores for models from the same provider (Gemini and Gemma). Results are reported in \Cref{tab:pipeline_validation_bias}. Overall, we find no substantial evidence of a pro-Google grading bias. For both tasks with textual outputs and tasks with image outputs, agreement is comparable between Google and non-Google models. The false positive rate is instead higher for non-Google models, meaning that incorrect answers are more likely to be promoted to "correct" for non-Google outputs than for Google model generations.

\paragraph{Evaluated Models Choice and Configuration}
We evaluate a diverse set of state-of-the-art LLMs, VLMs and image generation models, covering a range of sizes and capabilities. The following model families are included: \texttt{Qwen3} and \texttt{Qwen3.5} \citep{qwen2025qwen3, qwen2026qwen35omni}, \texttt{GLM 4.7, 5, 5.1, 5.2} \citep{zeng2026glm5}, \texttt{Kimi K2.5} and \texttt{2.6} \citep{kimi2026k25}, \texttt{DeepSeek V4} \citep{deepseek2026v4}, \texttt{Gemma-4} \cite{google2026gemma4}, and \texttt{GPT-OSS} \citep{agarwal2025gptoss}. For frontier models, we consider Google's \texttt{Gemini-2.5, 3, 3.1} \citep{comanici2025gemini25, google2025gemini3, google2026gemini31} and OpenAI \texttt{GPT-5, 5.2, 5.4, 5.5} \citep{openai2025gpt5, openai2026gpt54, openai2026gpt55}. When possible, we include multiple sizes of the same model to analyze the impact of scaling model size. VLMs are tested on both multi-to-text and text-only questions, while LLMs are evaluated on text-only questions. The evaluation is repeated $k=4$ times for tasks with textual output, and only once for image generation tasks, due to their significantly higher cost.

All models are evaluated with \texttt{thinking} mode enabled, and \texttt{medium} effort when these options are available, with the maximum number of output tokens set to 32{,}768. Unless stated otherwise, no model has access to tools. Generation requests are retried at most five times in case of failure, and are counted as incorrect if the model fails to produce any response. All open-weight LLMs and VLMs were accessed through EPFL Research Computing Platform (RCP) AI-Inference-as-a-Service (AIaaS) which deploys and operates AI models, using using vLLM \citep{kwon2023efficient} as backbone and exposing OpenAI-comapatible APIs.

For image generation tasks we evaluate different versions of the \texttt{GPT-Image} and \texttt{Gemini-Image} models, which are specifically designed for generative image tasks. Due to technical constraints, no open-weight image generation models are included.

A detailed list of the models along with their configuration is provided in \autoref{appendix:models_info}, where the reported open models pricing refers to the offering by RCP AIaaS.

\subsection{Main Results}
\colorlet{takeawaybg}{blue!10}
\newtcolorbox{takeawaysbox}{
    enhanced,
    colback=takeawaybg,
    colframe=black,
    boxrule=1.2pt,
    arc=6pt, 
    auto outer arc,
    top=10pt,
    bottom=8pt,
    left=6pt,
    right=6pt,
    fontupper=\normalsize,
    title=Takeaways,
    coltitle=white,
    fonttitle=\bfseries,
    attach boxed title to top left={xshift=12pt, yshift=-8pt},
    boxed title style={
        colback=black,
        colframe=black,
        size=small,
        arc=3pt, 
        bottom=1pt, top=1pt, left=4pt, right=4pt
    }
}

\begin{takeawaysbox}
\begin{itemize}[leftmargin=1.5em, topsep=2pt, itemsep=0pt]
    \item \textbf{Closed source models are more accurate} than open-weight ones, even at the same level of capabilities on other established benchmarks.
    \item \textbf{Open models are more cost-effective}: GLM-5.2, Qwen3.5 and DeepSeek-V4 perform better than frontier models \textit{with the same evaluation cost}.
    \item \textbf{Tool-use is not always beneficial}: it reduces token-use but sometimes decreases accuracy.
\end{itemize}
\end{takeawaysbox}

\paragraph{Leaderboard} \label{subsec:leaderboard}
The complete evaluation results are reported in \autoref{tab:llm-vlm-leaderboard}, \autoref{tab:image-gen-leaderboard}. We track performance by question type, reporting average accuracy (\textit{mean@k}) and success within \textit{k} attempts (\textit{pass@k}, defined as the probability of obtaining at least one correct solution out of \textit{k} independent attempts), as well as average inference cost and the number of generated tokens.
\begin{table}[t]
\small
\setlength{\tabcolsep}{4pt}
\caption{LLM and VLM model results by evaluation subset. Accuracy and output length are reported as $\texttt{mean}_{\pm \texttt{stderr}}$. Evaluation cost in reported in USD-per-100-samples.}
\label{tab:llm-vlm-leaderboard}
\begin{tabular}{lrrrrrrrr}
\toprule
 & \multicolumn{4}{c}{\textbf{Text-only}} & \multicolumn{4}{c}{\textbf{Multi-to-text}} \\
\cmidrule(lr){2-5} \cmidrule(lr){6-9}
Model & mean@4 & pass@4 & out-tks & cost & mean@4 & pass@4 & out-tks & cost \\
\midrule
\multicolumn{9}{l}{\textbf{Text-only models}} \\
GLM-4.7 & \cellcolor{green!54!red!25!white}{$61.3_{ \pm 4.1}$} & \cellcolor{green!63!red!25!white}{$75.9_{ \pm 4.1}$} & \cellcolor{green!42!red!25!white}{$5346_{ \pm 419}$} & \cellcolor{green!66!red!25!white}{0.91} & - & - & - & - \\
GLM-5 & \cellcolor{green!58!red!25!white}{$64.4_{ \pm 4}$} & \cellcolor{green!66!red!25!white}{$77.8_{ \pm 4}$} & \cellcolor{green!61!red!25!white}{$3404_{ \pm 216}$} & \cellcolor{green!62!red!25!white}{1.11} & - & - & - & - \\
GLM-5.1 & \cellcolor{green!62!red!25!white}{$67.4_{ \pm 4.1}$} & \cellcolor{green!65!red!25!white}{$76.9_{ \pm 4.1}$} & \cellcolor{green!65!red!25!white}{$2948_{ \pm 250}$} & \cellcolor{green!71!red!25!white}{0.69} & - & - & - & - \\
GLM-5.2 & \cellcolor{green!71!red!25!white}{$73.8_{ \pm 3.6}$} & \cellcolor{green!76!red!25!white}{$84.3_{ \pm 3.5}$} & \cellcolor{green!43!red!25!white}{$5229_{ \pm 486}$} & \cellcolor{green!62!red!25!white}{1.09} & - & - & - & - \\
DeepSeek-V4-Flash & \cellcolor{green!63!red!25!white}{$67.6_{ \pm 3.9}$} & \cellcolor{green!69!red!25!white}{$79.6_{ \pm 3.9}$} & \cellcolor{green!59!red!25!white}{$3648_{ \pm 391}$} & \cellcolor{green!79!red!25!white}{0.29} & - & - & - & - \\
DeepSeek-V4-Pro & \cellcolor{green!66!red!25!white}{$70.4_{ \pm 3.8}$} & \cellcolor{green!70!red!25!white}{$80.6_{ \pm 3.8}$} & \cellcolor{green!72!red!25!white}{$2232_{ \pm 201}$} & \cellcolor{green!72!red!25!white}{0.63} & - & - & - & - \\
GPT-oss-20b & \cellcolor{green!42!red!25!white}{$52.3_{ \pm 4.1}$} & \cellcolor{green!51!red!25!white}{$67.6_{ \pm 4.5}$} & \cellcolor{green!69!red!25!white}{$2611_{ \pm 317}$} & \cellcolor{green!85!red!25!white}{0.02} & - & - & - & - \\
GPT-oss-120b & \cellcolor{green!53!red!25!white}{$60.2_{ \pm 4.1}$} & \cellcolor{green!58!red!25!white}{$72.2_{ \pm 4.3}$} & \cellcolor{green!79!red!25!white}{$1555_{ \pm 204}$} & \cellcolor{green!85!red!25!white}{0.02} & - & - & - & - \\
Qwen3-Next-80B-A3B & \cellcolor{green!52!red!25!white}{$59.5_{ \pm 4.2}$} & \cellcolor{green!58!red!25!white}{$72.2_{ \pm 4.3}$} & \cellcolor{green!31!red!25!white}{$6460_{ \pm 444}$} & \cellcolor{green!80!red!25!white}{0.27} & - & - & - & - \\
\midrule
\multicolumn{9}{l}{\textbf{VLMs}} \\
Kimi-K2.5 & \cellcolor{green!64!red!25!white}{$68.3_{ \pm 3.8}$} & \cellcolor{green!70!red!25!white}{$80.6_{ \pm 3.8}$} & \cellcolor{green!54!red!25!white}{$4163_{ \pm 336}$} & \cellcolor{green!67!red!25!white}{0.86} & \cellcolor{green!50!red!25!white}{$42.4_{ \pm 6.8}$} & \cellcolor{green!45!red!25!white}{$53.5_{ \pm 7.7}$} & \cellcolor{green!34!red!25!white}{$4994_{ \pm 690}$} & \cellcolor{green!69!red!25!white}{1.07} \\
Kimi-K2.6 & \cellcolor{green!59!red!25!white}{$65_{ \pm 3.9}$} & \cellcolor{green!68!red!25!white}{$78.7_{ \pm 4}$} & \cellcolor{green!57!red!25!white}{$3797_{ \pm 266}$} & \cellcolor{green!70!red!25!white}{0.7} & \cellcolor{green!49!red!25!white}{$41.9_{ \pm 6.8}$} & \cellcolor{green!45!red!25!white}{$53.5_{ \pm 7.7}$} & \cellcolor{green!40!red!25!white}{$4501_{ \pm 527}$} & \cellcolor{green!72!red!25!white}{0.87} \\
Gemma-4-E2B-it & \cellcolor{green!15!red!25!white}{$32.4_{ \pm 4}$} & \cellcolor{green!15!red!25!white}{$42.6_{ \pm 4.8}$} & \cellcolor{green!79!red!25!white}{$1533_{ \pm 127}$} & \cellcolor{green!85!red!25!white}{0.01} & \cellcolor{green!15!red!25!white}{$18_{ \pm 4.7}$} & \cellcolor{green!15!red!25!white}{$30.2_{ \pm 7.1}$} & \cellcolor{green!85!red!25!white}{$1049_{ \pm 139}$} & \cellcolor{green!85!red!25!white}{0.01} \\
Gemma-4-E4B-it & \cellcolor{green!27!red!25!white}{$41.4_{ \pm 4}$} & \cellcolor{green!37!red!25!white}{$57.4_{ \pm 4.8}$} & \cellcolor{green!79!red!25!white}{$1601_{ \pm 147}$} & \cellcolor{green!85!red!25!white}{0.01} & \cellcolor{green!22!red!25!white}{$22.7_{ \pm 5.8}$} & \cellcolor{green!18!red!25!white}{$32.6_{ \pm 7.2}$} & \cellcolor{green!81!red!25!white}{$1335_{ \pm 287}$} & \cellcolor{green!85!red!25!white}{0.01} \\
Gemma-4-26B-A4B-it & \cellcolor{green!56!red!25!white}{$62.5_{ \pm 4.2}$} & \cellcolor{green!58!red!25!white}{$72.2_{ \pm 4.3}$} & \cellcolor{green!50!red!25!white}{$4541_{ \pm 341}$} & \cellcolor{green!84!red!25!white}{0.06} & \cellcolor{green!48!red!25!white}{$40.7_{ \pm 6.9}$} & \cellcolor{green!45!red!25!white}{$53.5_{ \pm 7.7}$} & \cellcolor{green!52!red!25!white}{$3587_{ \pm 578}$} & \cellcolor{green!84!red!25!white}{0.05} \\
Qwen3-VL-30B-A3B & \cellcolor{green!38!red!25!white}{$49.5_{ \pm 4.1}$} & \cellcolor{green!50!red!25!white}{$66.7_{ \pm 4.6}$} & \cellcolor{green!58!red!25!white}{$3673_{ \pm 256}$} & \cellcolor{green!84!red!25!white}{0.07} & \cellcolor{green!25!red!25!white}{$25_{ \pm 6}$} & \cellcolor{green!24!red!25!white}{$37.2_{ \pm 7.5}$} & \cellcolor{green!62!red!25!white}{$2796_{ \pm 538}$} & \cellcolor{green!84!red!25!white}{0.06} \\
Qwen3-VL-235B-A22B & \cellcolor{green!50!red!25!white}{$58.3_{ \pm 4.2}$} & \cellcolor{green!57!red!25!white}{$71.3_{ \pm 4.4}$} & \cellcolor{green!61!red!25!white}{$3435_{ \pm 222}$} & \cellcolor{green!78!red!25!white}{0.32} & \cellcolor{green!35!red!25!white}{$32_{ \pm 6.8}$} & \cellcolor{green!27!red!25!white}{$39.5_{ \pm 7.5}$} & \cellcolor{green!65!red!25!white}{$2604_{ \pm 351}$} & \cellcolor{green!81!red!25!white}{0.26} \\
Qwen3.5-35B-A3B & \cellcolor{green!58!red!25!white}{$63.9_{ \pm 4.2}$} & \cellcolor{green!61!red!25!white}{$74.1_{ \pm 4.2}$} & \cellcolor{green!23!red!25!white}{$7278_{ \pm 374}$} & \cellcolor{green!83!red!25!white}{0.11} & \cellcolor{green!57!red!25!white}{$47.1_{ \pm 6.5}$} & \cellcolor{green!67!red!25!white}{$69.8_{ \pm 7.1}$} & \cellcolor{green!15!red!25!white}{$6492_{ \pm 643}$} & \cellcolor{green!84!red!25!white}{0.1} \\
Qwen3.5-122B-A10B & \cellcolor{green!58!red!25!white}{$63.9_{ \pm 4.2}$} & \cellcolor{green!62!red!25!white}{$75_{ \pm 4.2}$} & \cellcolor{green!39!red!25!white}{$5637_{ \pm 284}$} & \cellcolor{green!80!red!25!white}{0.25} & \cellcolor{green!59!red!25!white}{$48.8_{ \pm 6.9}$} & \cellcolor{green!58!red!25!white}{$62.8_{ \pm 7.5}$} & \cellcolor{green!55!red!25!white}{$3350_{ \pm 353}$} & \cellcolor{green!83!red!25!white}{0.16} \\
Qwen3.5-397B-A17B & \cellcolor{green!67!red!25!white}{$71.1_{ \pm 3.9}$} & \cellcolor{green!72!red!25!white}{$81.5_{ \pm 3.8}$} & \cellcolor{green!37!red!25!white}{$5849_{ \pm 288}$} & \cellcolor{green!66!red!25!white}{0.89} & \cellcolor{green!58!red!25!white}{$48.3_{ \pm 7.2}$} & \cellcolor{green!52!red!25!white}{$58.1_{ \pm 7.6}$} & \cellcolor{green!28!red!25!white}{$5479_{ \pm 502}$} & \cellcolor{green!73!red!25!white}{0.85} \\
Gemini-2.5-flash & \cellcolor{green!46!red!25!white}{$55.3_{ \pm 4.1}$} & \cellcolor{green!55!red!25!white}{$70.4_{ \pm 4.4}$} & \cellcolor{green!56!red!25!white}{$3884_{ \pm 513}$} & \cellcolor{green!65!red!25!white}{0.97} & \cellcolor{green!35!red!25!white}{$32_{ \pm 6.5}$} & \cellcolor{green!30!red!25!white}{$41.9_{ \pm 7.6}$} & \cellcolor{green!55!red!25!white}{$3383_{ \pm 997}$} & \cellcolor{green!73!red!25!white}{0.85} \\
Gemini-2.5-pro & \cellcolor{green!53!red!25!white}{$60.4_{ \pm 4}$} & \cellcolor{green!65!red!25!white}{$76.9_{ \pm 4.1}$} & \cellcolor{green!66!red!25!white}{$2905_{ \pm 311}$} & \cellcolor{green!24!red!25!white}{2.92} & \cellcolor{green!41!red!25!white}{$36_{ \pm 6.8}$} & \cellcolor{green!33!red!25!white}{$44.2_{ \pm 7.7}$} & \cellcolor{green!56!red!25!white}{$3283_{ \pm 748}$} & \cellcolor{green!37!red!25!white}{3.32} \\
Gemini-3-flash & \cellcolor{green!74!red!25!white}{$76.2_{ \pm 3.7}$} & \cellcolor{green!74!red!25!white}{$83.3_{ \pm 3.6}$} & \cellcolor{green!15!red!25!white}{$8130_{ \pm 637}$} & \cellcolor{green!34!red!25!white}{2.45} & \cellcolor{green!82!red!25!white}{$64.5_{ \pm 6.2}$} & \cellcolor{green!85!red!25!white}{$83.7_{ \pm 5.7}$} & \cellcolor{green!60!red!25!white}{$2942_{ \pm 639}$} & \cellcolor{green!71!red!25!white}{0.94} \\
Gemini-3.1-flash-lite & \cellcolor{green!55!red!25!white}{$61.8_{ \pm 3.9}$} & \cellcolor{green!68!red!25!white}{$78.7_{ \pm 4}$} & \cellcolor{green!79!red!25!white}{$1537_{ \pm 166}$} & \cellcolor{green!80!red!25!white}{0.23} & \cellcolor{green!50!red!25!white}{$42.4_{ \pm 7.1}$} & \cellcolor{green!42!red!25!white}{$51.2_{ \pm 7.7}$} & \cellcolor{green!81!red!25!white}{$1335_{ \pm 112}$} & \cellcolor{green!82!red!25!white}{0.23} \\
Gemini-3.1-pro & \cellcolor{green!84!red!25!white}{$83.3_{ \pm 3.2}$} & \cellcolor{green!85!red!25!white}{$90.7_{ \pm 2.8}$} & \cellcolor{green!40!red!25!white}{$5550_{ \pm 645}$} & \cellcolor{green!15!red!25!white}{3.34} & \cellcolor{green!85!red!25!white}{$66.9_{ \pm 6.4}$} & \cellcolor{green!76!red!25!white}{$76.7_{ \pm 6.5}$} & \cellcolor{green!47!red!25!white}{$3959_{ \pm 858}$} & \cellcolor{green!49!red!25!white}{2.49} \\
GPT-5-mini & \cellcolor{green!63!red!25!white}{$67.6_{ \pm 4}$} & \cellcolor{green!66!red!25!white}{$77.8_{ \pm 4}$} & \cellcolor{green!74!red!25!white}{$2032_{ \pm 225}$} & \cellcolor{green!77!red!25!white}{0.41} & \cellcolor{green!37!red!25!white}{$33.1_{ \pm 6.5}$} & \cellcolor{green!33!red!25!white}{$44.2_{ \pm 7.7}$} & \cellcolor{green!80!red!25!white}{$1437_{ \pm 221}$} & \cellcolor{green!81!red!25!white}{0.3} \\
GPT-5 & \cellcolor{green!67!red!25!white}{$70.6_{ \pm 3.8}$} & \cellcolor{green!73!red!25!white}{$82.4_{ \pm 3.7}$} & \cellcolor{green!64!red!25!white}{$3104_{ \pm 329}$} & \cellcolor{green!20!red!25!white}{3.12} & \cellcolor{green!43!red!25!white}{$37.8_{ \pm 6.7}$} & \cellcolor{green!42!red!25!white}{$51.2_{ \pm 7.7}$} & \cellcolor{green!69!red!25!white}{$2279_{ \pm 282}$} & \cellcolor{green!51!red!25!white}{2.35} \\
GPT-5.2 & \cellcolor{green!67!red!25!white}{$70.6_{ \pm 3.7}$} & \cellcolor{green!74!red!25!white}{$83.3_{ \pm 3.6}$} & \cellcolor{green!83!red!25!white}{$1165_{ \pm 169}$} & \cellcolor{green!50!red!25!white}{1.66} & \cellcolor{green!61!red!25!white}{$50_{ \pm 7}$} & \cellcolor{green!55!red!25!white}{$60.5_{ \pm 7.5}$} & \cellcolor{green!85!red!25!white}{$1022_{ \pm 221}$} & \cellcolor{green!63!red!25!white}{1.54} \\
GPT-5.4-nano & \cellcolor{green!55!red!25!white}{$62_{ \pm 4}$} & \cellcolor{green!63!red!25!white}{$75.9_{ \pm 4.1}$} & \cellcolor{green!77!red!25!white}{$1734_{ \pm 276}$} & \cellcolor{green!81!red!25!white}{0.22} & \cellcolor{green!25!red!25!white}{$25_{ \pm 6.2}$} & \cellcolor{green!18!red!25!white}{$32.6_{ \pm 7.2}$} & \cellcolor{green!81!red!25!white}{$1369_{ \pm 367}$} & \cellcolor{green!82!red!25!white}{0.18} \\
GPT-5.4-mini & \cellcolor{green!56!red!25!white}{$63_{ \pm 4}$} & \cellcolor{green!66!red!25!white}{$77.8_{ \pm 4}$} & \cellcolor{green!73!red!25!white}{$2205_{ \pm 389}$} & \cellcolor{green!64!red!25!white}{1} & \cellcolor{green!54!red!25!white}{$45.3_{ \pm 6.7}$} & \cellcolor{green!58!red!25!white}{$62.8_{ \pm 7.5}$} & \cellcolor{green!70!red!25!white}{$2224_{ \pm 847}$} & \cellcolor{green!70!red!25!white}{1.05} \\
GPT-5.4 & \cellcolor{green!78!red!25!white}{$78.9_{ \pm 3.2}$} & \cellcolor{green!85!red!25!white}{$90.7_{ \pm 2.8}$} & \cellcolor{green!82!red!25!white}{$1299_{ \pm 181}$} & \cellcolor{green!44!red!25!white}{1.98} & \cellcolor{green!72!red!25!white}{$58.1_{ \pm 6.7}$} & \cellcolor{green!73!red!25!white}{$74.4_{ \pm 6.7}$} & \cellcolor{green!80!red!25!white}{$1420_{ \pm 286}$} & \cellcolor{green!52!red!25!white}{2.29} \\
GPT-5.5 & \cellcolor{green!85!red!25!white}{$84_{ \pm 3.2}$} & \cellcolor{green!84!red!25!white}{$89.8_{ \pm 2.9}$} & \cellcolor{green!85!red!25!white}{$945_{ \pm 110}$} & \cellcolor{green!24!red!25!white}{2.89} & \cellcolor{green!73!red!25!white}{$58.7_{ \pm 6.5}$} & \cellcolor{green!73!red!25!white}{$74.4_{ \pm 6.7}$} & \cellcolor{green!79!red!25!white}{$1491_{ \pm 336}$} & \cellcolor{green!15!red!25!white}{4.8} \\
\bottomrule
\end{tabular}
\end{table}

\paragraph{Overall Performance}
\texttt{Gemini-3.1-Pro} achieves the strongest overall performance, with an accuracy of $\approx$83.3\% on textual problems and $\approx$66.9\% on multimodal problems \autoref{tab:llm-vlm-leaderboard}. While \texttt{GPT-5.5} also attains very high accuracy on textual problems ($\approx$84\%), its performance does not transfer to visual questions ($\approx$58.7\%). The strongest open-weight models remain competitive mainly on text-only problems, but lag behind in the multimodal setting. \texttt{GLM-5.2} leads among open-weights models with a text-only accuracy of $\approx$74\%. \texttt{Qwen3.5-397B} also demonstrates strong performance ($\approx$70\% on text, $\approx$48\% on multimodal), surpassing much larger models such as \texttt{Kimi-K2.6} and \texttt{DeepSeek-V4-Pro}. This result might come as a consequence of longer reasoning, as suggested by the high average number of output tokens.
Overall we observe substantial improvements across successive iterations of the same model families in the different iterations of the same model families (e.g., from \texttt{Gemini-2.5} to \texttt{Gemini-3}), indicating that the latest generation of models is significantly more robust than models from late 2025.

For image-generation models~\autoref{tab:image-gen-leaderboard}) \texttt{Gemini-3-Pro-Image} attains highest accuracy (54.8\%), but at a substantially higher inference cost than its OpenAI counterpart. \texttt{GPT-Image-2} achieves slightly lower accuracy (51.2\%) while being approximately 4$\times$ cheaper. A detailed view of the failures distribution across the different problems is provided in \autoref{appendix:full_accuracy}.

\begin{figure}
    \centering
    \includegraphics[width=\linewidth]{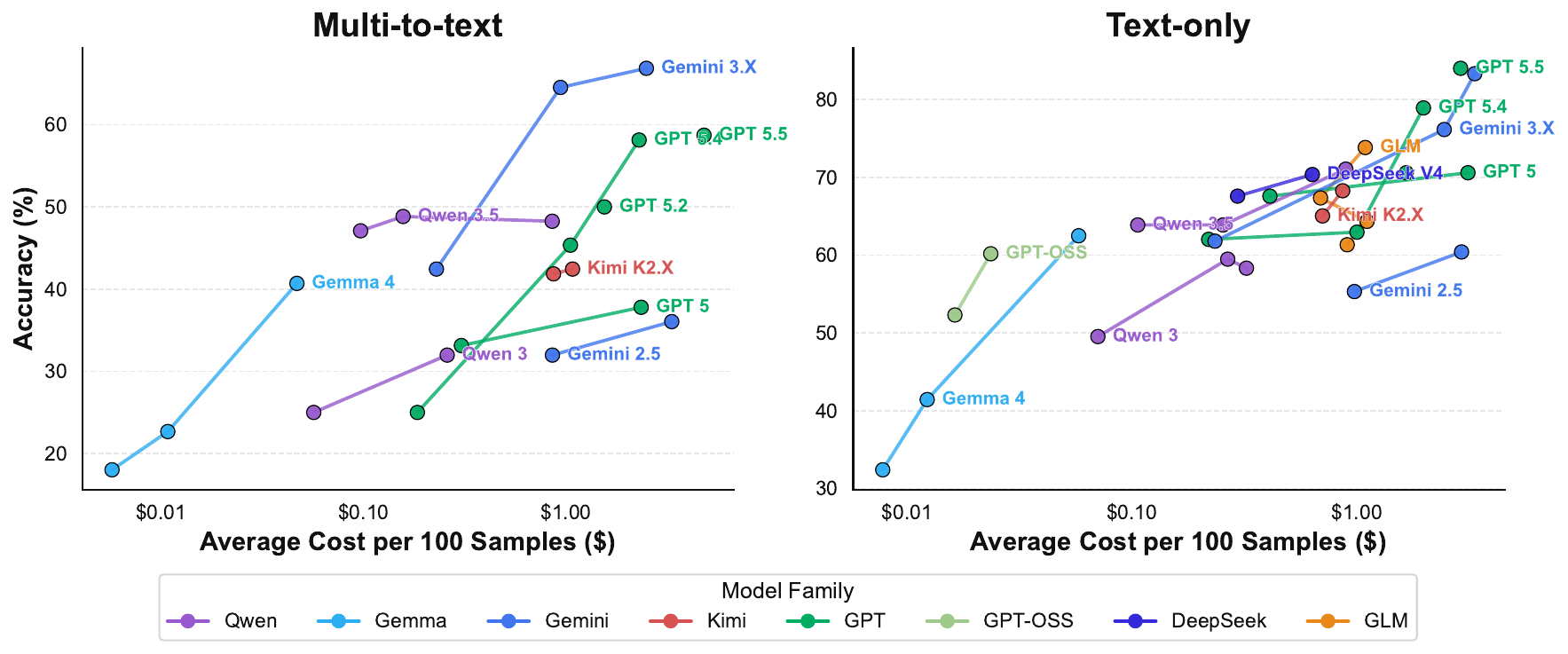}
    \caption{
    Accuracy on \bench\ vs. average cost for 100 samples. Colors distinguish model families; models of the same version but different sizes are connected.
    }
    \label{fig:acc_vs_cost}
\end{figure}

\paragraph{Cost-Performance Trade-off}
We further analyze the relation between each model's accuracy on \bench\ and its average inference cost. As shown in \autoref{fig:acc_vs_cost}, the rate of improvement in accuracy slowly decreases with higher cost. In text-only tasks in particular, we observe that a 10\% accuracy gain, from 60\% to 70\%, requires at least an order-of-magnitude increase in inference cost (approximately 10$\times$).
Few open models, in paricular \texttt{DeepSeek-V4}, \texttt{Qwen3.5}, and \texttt{GLM-5.2} demonstrate particularly cost-efficient performance, outperforming the frontier alternatives at similar cost levels (such as \texttt{Gemini-3.1-Flash-Lite} and \texttt{GPT-5.4-mini}).
Among frontier models, the \texttt{Gemini-3} family displays an overall better accuracy-cost trade-off, especially in the multimodal domain, strengthening the evidence of its greater performance when compared to \texttt{GPT-5.4} and \texttt{5.5}. 

\paragraph{Alignment with General Intelligence Assessment}
We further compare \bench\ accuracy with external general intelligence scores, focusing on the Artificial Analysis Intelligence Index (AAII) \citep{artificialanalysis_methodology_2026}. This composite benchmark aggregates ten established text-only evaluations that span diverse capabilities, including Humanity Last Exam \citep{phan2025humanitys_last_exam}, IFBench \citep{pyatkin2025ifbench}, and TerminalBench \citep{merrill2026terminal_bench}. As shown in \autoref{fig:acc_vs_aaii_text}a, we observe a positive relationship between Intelligence Index scores and performance \bench, with models that perform well on broad reasoning, coding, and mathematical evaluations also scoring highly on \bench\footnote{For some models, an AAII score is not available at exactly \texttt{medium} thinking effort. In these cases, we use the best available approximation: the average of \texttt{low} and \texttt{high} when both are reported, or the single reasoning-enabled score when no \texttt{thinking\_effort} option is provided.}. This supports the view that \bench measures general reasoning ability rather than an capturing an unrelated set of adversarial prompts.

However, this relation differs between open- and closed-weight models. Open models tend to perform worse on \bench\ than closed models with comparable AAII scores. One possible explanation is that broad public benchmarks may overestimate robustness on underrepresented tasks, especially when models have been optimized for widely used evaluations that emphasize specific skills rather than global robustness.

\begin{table}
    \centering
    \begin{minipage}[t]{0.48\textwidth}
        \vspace{0pt}
\captionof{table}{Image-generation model results. Accuracy ($\texttt{mean}_{\pm \texttt{stderr}}$) is shown as a percentage, generation cost as USD-per-100-samples.}
        \centering
\begin{tabular}{lrr}
\toprule
 & \multicolumn{2}{c}{\textbf{Image-gen}} \\
\cmidrule(lr){2-3}
Model & mean@1 & cost \\
\midrule
\multicolumn{3}{l}{\textbf{Image-generation models}} \\
Gemini-2.5-flash-image & \cellcolor{green!22!red!25!white}{$22.6_{ \pm 4.6}$} & \cellcolor{green!72!red!25!white}{4.14} \\
Gemini-3-pro-image & \cellcolor{green!85!red!25!white}{$54.8_{ \pm 5.5}$} & \cellcolor{green!15!red!25!white}{19.23} \\
Gemini-3.1-flash-image & \cellcolor{green!83!red!25!white}{$53.6_{ \pm 5.5}$} & \cellcolor{green!52!red!25!white}{9.56} \\
GPT-image-1-mini & \cellcolor{green!15!red!25!white}{$19_{ \pm 4.3}$} & \cellcolor{green!85!red!25!white}{0.85} \\
GPT-image-1.5 & \cellcolor{green!57!red!25!white}{$40.5_{ \pm 5.4}$} & \cellcolor{green!71!red!25!white}{4.59} \\
GPT-image-2 & \cellcolor{green!78!red!25!white}{$51.2_{ \pm 5.5}$} & \cellcolor{green!68!red!25!white}{5.26} \\
\bottomrule
\end{tabular}
    \label{tab:image-gen-leaderboard}
    \end{minipage}
    \hfill
    \begin{minipage}[t]{0.48\textwidth}
        \vspace{0pt}
        \centering
            \captionof{table}{Delta in performance and tokens, when enabling code execution for text-only problems.}
        \centering
\footnotesize
\setlength{\tabcolsep}{4pt}
\label{tab:tool-use-comparison}
\begin{tabular}{lrr}
\toprule
 & \multicolumn{2}{c}{\textbf{Text-only}} \\
\cmidrule(lr){2-3}
Model & acc & out-tks \\
\midrule
Gemini-3.1-pro & 86.11 {\scriptsize\textcolor{green!70!black}{(+2.78)}} & 5352 {\scriptsize\textcolor{green!70!black}{(-199)}} \\
GLM-5.2 & 75 {\scriptsize\textcolor{green!70!black}{(+1.16)}} & 3518 {\scriptsize\textcolor{green!70!black}{(-1712)}} \\
GLM-5.1 & 75 {\scriptsize\textcolor{green!70!black}{(+7.64)}} & 2614 {\scriptsize\textcolor{green!70!black}{(-334)}} \\
GPT-5.4 & 73.61 {\scriptsize\textcolor{red!70!black}{(-5.32)}} & 848 {\scriptsize\textcolor{green!70!black}{(-451)}} \\
Kimi-K2.6 & 71.3 {\scriptsize\textcolor{green!70!black}{(+6.25)}} & 7112 {\scriptsize\textcolor{red!70!black}{(+3314)}} \\
Gemini-3.1-flash-lite & 70.83 {\scriptsize\textcolor{green!70!black}{(+9.03)}} & 983 {\scriptsize\textcolor{green!70!black}{(-554)}} \\
Qwen3.5-397B-A17B & 68.98 {\scriptsize\textcolor{red!70!black}{(-2.08)}} & 3068 {\scriptsize\textcolor{green!70!black}{(-2781)}} \\
Kimi-K2.5 & 68.06 {\scriptsize\textcolor{red!70!black}{(-0.23)}} & 4938 {\scriptsize\textcolor{red!70!black}{(+775)}} \\
GPT-oss-120b & 67.13 {\scriptsize\textcolor{green!70!black}{(+6.94)}} & 1462 {\scriptsize\textcolor{green!70!black}{(-94)}} \\
GPT-5.4-mini & 66.67 {\scriptsize\textcolor{green!70!black}{(+3.7)}} & 1507 {\scriptsize\textcolor{green!70!black}{(-698)}} \\
Qwen3-VL-30B-A3B & 56.94 {\scriptsize\textcolor{green!70!black}{(+7.41)}} & 3630 {\scriptsize\textcolor{green!70!black}{(-43)}} \\
\bottomrule
\end{tabular}
        \label{tab:tool-use-comparison}
    \end{minipage}
\end{table}

\paragraph{Tool Use Impact on Reasoning}
Some of the most challenging tasks for LLMs often involve meeting hard constraints, counting and performing string operations. Because of this, the use of external tools, particularly code execution environments, can substantially improve performance.
Therefore, we repeat the evaluation for a subset of models, equipping them with a Python execution environment and allowing up to five tool calls per question. Results in \autoref{tab:tool-use-comparison} show mixed effects, with some models like \texttt{Gemini-3.1-Flash} achieving notably higher accuracy with reduced token consumption, whereas others, such as \texttt{GPT-5.4} and \texttt{Qwen3.5-397B}, lose performance. A qualitative evaluation of these failures indicates that, despite generating correct code snippets, models often mishandle inputs, for example, when copy-pasting an input string.

\begin{figure}[htbp!]
    \centering
    \includegraphics[width=\linewidth]{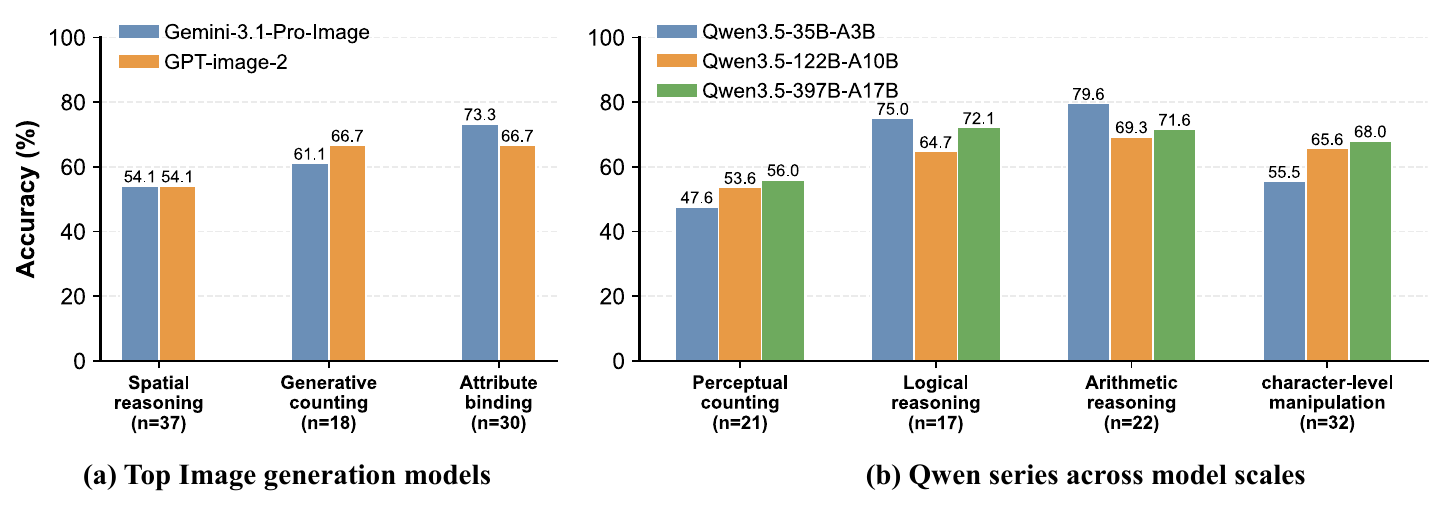}
    \caption{Performance comparison of leading image generation and VLM models on the largest-sample subtasks within each model type.}
    \label{fig:task_analysis}
\end{figure}
\subsection{Results Analysis by Task Taxonomy}

We analyze model performance across the defined taxonomy to better understand  their strengths and limitations, with the full results reported in  \autoref{tab:analyses_category} and \autoref{tab:analyses_category-2}. Scores are computed using mean@4 for all models  except image-generation models. For the 15 questions annotated with multiple subtask labels, we further validate their fine-grained task scores. Since different question formats contain different numbers of subtask categories, we mainly conduct the 
analysis within each input-output format. Several interesting findings emerge. 


First, \textbf{\emph{fine-grained visual perception remains a major bottleneck, with models consistently struggling on attribute and pattern recognition and perceptual counting.}} Even the strongest models obtain only 41.67\%  and 57.14\% on these two subtasks, which contain 6 and 21 questions, respectively. For perceptual counting, many models achieve only around 30--40\% accuracy. 

Second, \textbf{\emph{different model families exhibit complementary strengths rather than a single dominant pattern across all subtask categories.}} Here, we compare among the best-performing model from each family. For image-generation models, Gemini  and GPT achieve comparable overall performance, but exhibit complementary strengths:  Gemini performs better on abstract reasoning winning 3 more questions, whereas GPT is stronger in language and knowledge~\autoref{fig:task_analysis}. A similar pattern appears among text-only models. GPT-OSS remains competitive on math-related tasks, but lags behind others on all three language-and-knowledge subtasks. DeepSeek-V4 performs particularly well on character-level manipulation, reaching 75\% accuracy. Among VLMs, the GPT and Gemini series form the leading group. GPT trails Gemini on  object-centric tasks, but slightly outperforms it on the other two major categories. Although Qwen and Kimi sometimes perform strongly on individual subtasks, such as Kimi's 83\% accuracy on arithmetic reasoning, their performance is less consistent, with noticeable gaps from the leading models in other cases, including a 10-point gap on character-level manipulation.

Third, \textbf{\emph{model size scaling does not consistently improve performance across all subtask types.}} We compare different model sizes within the same model family and observe  several cases where larger models underperform smaller variants. For example, in 
the Gemma-4 series, the 26B model improves on most subtasks but performs about 8\%  worse than the E2B model on irrelevant-context robustness. Similarly, in the Qwen3.5 series, the 122B model shows a 10--14\% drop over the 35B model on several 
abstract-reasoning subtasks, including logical reasoning, geometric and graph  reasoning, and arithmetic reasoning. Although the 397B model recovers part of this drop, it still does not surpass the 35B model on these subtasks.

\section{Conclusion}
In this work, we present \bench, a dataset of 235 questions that remain challenging for modern models but straightforward for humans. We introduce a fine-grained taxonomy, curated reference solutions, and an automatic grading pipeline to evaluate diverse VLMs, LLMs, and image-generation models. Our analyses show that closed-source frontier models outperform open-weight models, while open-weight models are often more cost-effective. We further find that scaling model size or enabling tool use does not consistently improve performance. Finally, taxonomy-level analysis reveals several distinctive patterns in model failure modes.

\textbf{Limitations.}
While \bench provides useful insights into model failure modes, several limitations remain. First, the dataset is relatively modest in size and imbalanced across sub-tasks. Although sufficient for our analyses, increasing the number of samples per sub-task would strengthen the statistical reliability of the results. Second, because the questions were primarily created by students to challenge two frontier models, they may be biased toward those models' specific weaknesses and may miss other types of reasoning failures. Finally, although \bench questions are intended to be easy for humans, adding a human baseline would more directly quantify the gap between model and human performance.

\begin{ack}
This project was first proposed in the EPFL course Reasoning in Artificial Intelligence (MATH-700), taught by Emmanuel Abbé. We thank all students in the course for each proposing five questions to challenge frontier models at the time. We also thank Kevin Qiu, Amene Gafsi, Rafaila Galanopoulou, Szymon Sobczak, Julien David Laurendeau, and Zhipeng Xue for their early efforts to dataset cleaning. Finally, we thank the RCP team at EPFL for GPU and model support, and the Laboratory of Mathematical Data Science, which ran the course, for covering the associated costs.
\end{ack}

\bibliographystyle{unsrtnat}
\bibliography{bibliography}


\clearpage
\appendix

\section{Examples of problems} \label{appendix:examples}

To illustrate a practical example for the annotation and task taxonomy, we provide the following prompt from the dataset:
\begin{tcolorbox}[title={Example 1}, size=title,colback=gray!10, colframe=gray, boxrule=0.5pt,
  arc=2mm, left=2mm, right=2mm, top=1mm, bottom=1mm,,breakable ]
\noindent
\textbf{Prompt}: Can you read the words marked in this ``find a word'' game.
\begin{figure}[H]
    \centering
    \includegraphics[width=0.4\linewidth]{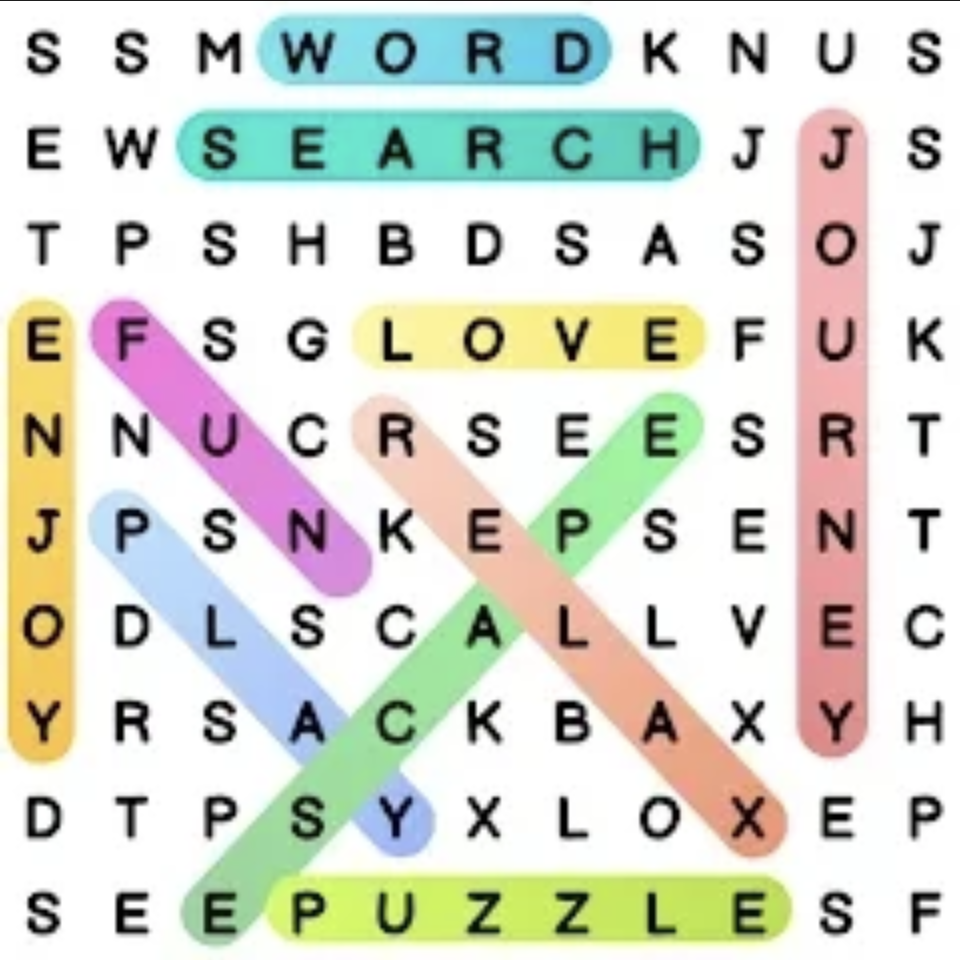}
    \label{fig:placeholder}
\end{figure}

\textbf{Solution}: The words marked in colors are: WORD, SEARCH, JOURNEY, LOVE, RELAX, ESCAPE, FUN, PLAY, PUZZLE, ENJOY. Any answer that does not contain one of these words, or contains an extra word not in this list, is considered wrong.\\

\textbf{Question type}: \texttt{multi-to-text}\\

\textbf{Task Type}: \toggle{object-centric} \\

\textbf{Sub-Task Type}: Attribute and pattern recognition.
\end{tcolorbox}

\begin{tcolorbox}[title={Example 2}, size=title,colback=gray!10, colframe=gray, boxrule=0.5pt,
  arc=2mm, left=2mm, right=2mm, top=1mm, bottom=1mm,,breakable ]
\noindent
\textbf{Prompt}: Given a directed weighted graph. What is the shortest path from A to B?
\begin{figure}[H]
    \centering
    \includegraphics[width=0.5\linewidth]{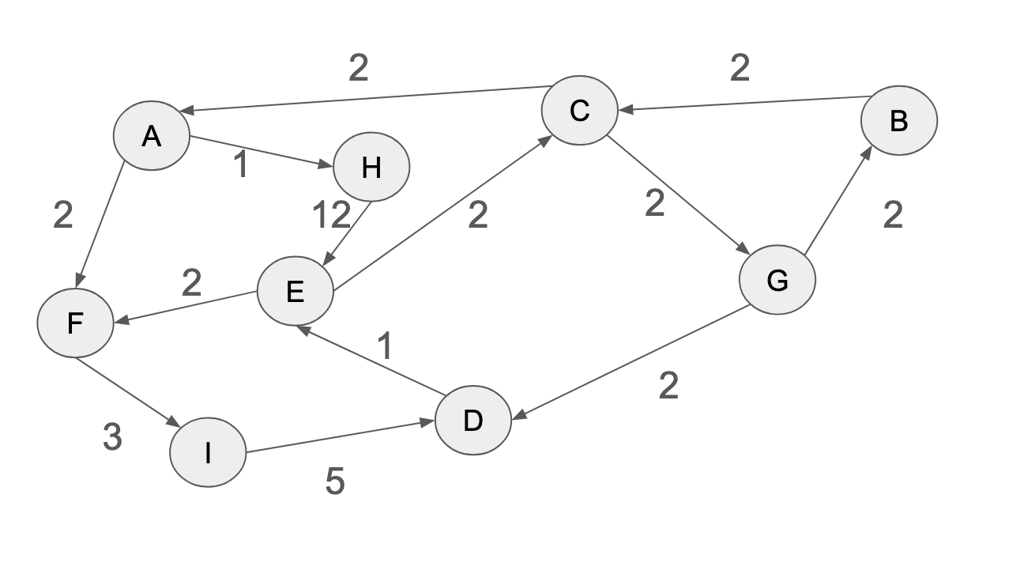}
    \label{fig:placeholder}
\end{figure}

\textbf{Solution}: The correct answer is a shortest directed path from A to B together with its total weight. An output is correct if and only if (1) every step follows an existing directed edge in the graph (respecting arrow direction) and (2) the total weight equals the minimum over all valid directed A→B paths.

The correct shortest path is A → F → I → D → E → C → G → B with total weight 17.

This is correct because each edge in the path exists with the shown direction and weights (A→F=2, F→I=3, I→D=5, D→E=1, E→C=2, C→G=2, G→B=2), summing to 17, and no other valid directed path from A to B has a smaller total weight.

A common incorrect behavior is ignoring edge directions (treating C→A as A→C or B→C as C→B) or inventing non-existent edges (e.g., claiming C→D or D→B), producing paths that are not actually reachable in the directed graph.\\

\textbf{Question type}: \texttt{multi-to-text}\\

\textbf{Task Type}: \toggle{Abstract reasoning} \\

\textbf{Sub-Task Type}: Geometric and graph reasoning.
\end{tcolorbox}


 \label{question_example}

We note that some prompts require multi-task categories, in this case we discuss further the failure mode depending on the model's answer. We provide an example below:
\begin{tcolorbox}[title={Multi category example}, size=title,colback=gray!10, colframe=gray, boxrule=0.5pt,
  arc=2mm, left=2mm, right=2mm, top=1mm, bottom=1mm,,breakable ]
\noindent
\textbf{Prompt}: Can you generate an image according to the problem description and solve the problem? Let ABCD be a square, and l be a line segment from B to a point on side AD. A is 5 cm from l and C is 7 cm from l. What is the area of ABCD? \\

\textbf{Question type}: \texttt{image-gen}\\

\textbf{Task Type}: \toggle{abstract reasoning}, \toggle{object-centric}\\

\textbf{Sub-Task Type}: Geometric and Graph reasoning, Attribute binding.\\

\textbf{Failure Mode}: If the model generates an image with incorrect properties, such as placing point B off side AD or producing square ABDC instead of ABCD, then it fails in attribute binding.
If the model fails to compute the area of ABCD, then it fails in geometric and graph reasoning.
\end{tcolorbox}
 \label{question_example_multi_category} 

\begin{table}[t]
\centering
\caption{We propose a taxonomy of capability dimensions, including object-centric, abstract reasoning, and language and knowledge, with detailed sub-task categories. For each question, we primarily identify its dominant task type for analysis; but still 15 questions have more than one requirement.}
\label{tab:dataset_category}
\setlength{\tabcolsep}{3pt}
\begin{adjustbox}{max width=\textwidth}
\newcolumntype{P}[1]{>{\normalfont\arraybackslash}p{#1}}
\begin{tabular}{lll P{0.43\textwidth}P{0.43\textwidth}}
\toprule
\bf Task Type & \bf Sub-Task type &  \bf ID & \normalfont{\textbf{Description}} &\textbf{ \normalfont{\textbf{Example}}} \\
\midrule

\multirow{5}{*}{\textbf{Object-centric}}
  & Attribute and pattern recognition & 1-1 & Perceive and ground objects/attributes/relations in visual inputs or required outputs (e.g., detection, attribute recognition). &
  \begin{minipage}[t]{\linewidth}
  \begin{itemize}[itemsep=0pt, topsep=0pt, parsep=0pt, partopsep=0pt, leftmargin=15pt]
      \item Given a picture, how many red and blue balls are in the picture?
  \end{itemize} \end{minipage}  \\ \addlinespace[8pt]
  & Spatial reasoning & 1-2 & The ability to perceive, represent, and manipulate spatial relations among objects. Also includes spatial transformations such as mirroring and flipping to an image or generated object. & \begin{minipage}[t]{\linewidth}
  \begin{itemize}[itemsep=0pt, topsep=0pt, parsep=0pt, partopsep=0pt, leftmargin=15pt]
      \item  Generate an image of a fully furnished living room. No furniture can touch the carpet. 
      \item Mirror the input image.
  \end{itemize}\end{minipage}
 \\ \addlinespace[8pt]
  & Generative counting & 1-3 & Enforce discrete quantity constraints during generation (e.g., generate exactly $k$ objects/parts). & \begin{minipage}[t]{\linewidth}
  \begin{itemize}[itemsep=0pt, topsep=0pt, parsep=0pt, partopsep=0pt, leftmargin=15pt]
      \item Make a figure that repeats a circle, a triangle, and a square 4 times. 
  \end{itemize}\end{minipage}
 \\ \addlinespace[6pt]
  & Perceptual counting & 1-4 & Perceptual counting is for counting visible objects in an image, like trees, chess pieces, dice, clouds. & \begin{minipage}[t]{\linewidth}
  \begin{itemize}[itemsep=0pt, topsep=0pt, parsep=0pt, partopsep=0pt, leftmargin=15pt]
      \item Given a picture, how many chess pieces are in the picture?
  \end{itemize}\end{minipage}
  \\ \addlinespace[8pt]
  & Attribute binding & 1-5 & Correctly associate object attributes such as color, shape, size, or identity with the corresponding objects, avoiding swaps or mismatches between objects and their properties. & \begin{minipage}[t]{\linewidth}
  \begin{itemize}[itemsep=0pt, topsep=0pt, parsep=0pt, partopsep=0pt, leftmargin=15pt]
      \item Generate an image with the writing "Google" where the two "g"s are red, "o"s are blue and green, "l" is red and "e" is yellow.
  \end{itemize} \end{minipage}
  \\ \addlinespace[8pt]
\midrule

\multirow{7}{*}{\textbf{Abstract reasoning}}
  & Logical reasoning & 2-1 &Task that requires multiple sequential steps of inference where the solution is derived through a directed chain of implications or valid state-changes.& 
 \begin{minipage}[t]{\linewidth} \begin{itemize}[itemsep=0pt, topsep=0pt, parsep=0pt, partopsep=0pt, leftmargin=15pt]
      \item If every square were a circle, and every circle were a triangle, what shape would a square be?
      \item Your task is to deduce the two steps of a security protocol and their correct order of application from the log file entries below. Then, use the protocol to find the correct output for the final two inputs. Log File Entries: \text{GOLD} $\rightarrow$ \text{KSPH}, 
\text{GLASS} $\rightarrow$ \text{XXFQL}, 
\text{STEEL} $\rightarrow$ \text{QJJYX}, 
\text{PASTA} $\rightarrow$ \text{FYXFU}, 
\text{IMAGE} $\rightarrow$ \text{NRFLJ}. 

Find the outputs for: 
\text{QUARTZ} $\rightarrow$ ?, 
\text{SILVER} $\rightarrow$ ?.
  \end{itemize} \end{minipage}
 \\ \addlinespace[8pt]
  & Arithmetic reasoning & 2-2 & Task that requires numerical operations, quantitative inference, working with arithmetic expression and reasoning under uncertainty using probabilistic rules.  &  \begin{minipage}[t]{\linewidth}\begin{itemize}[itemsep=0pt, topsep=0pt, parsep=0pt, partopsep=0pt, leftmargin=15pt]
      \item Find me three integers a, b, and c such that $\exp(9\log(a)) - \exp(18\log(c)/2) + (b^3)^3 = 0.$ Can you solve this with a, b and c being all non-zero?
      \item Give a smallest non-negative number which can be written as sum of cubes of two different pairs where each pair has different integers
      \item I draw a white marble from a bag with two marbles. What is the probability the other marble is also white?
  \end{itemize} \end{minipage}
\\ \addlinespace[8pt]
  & Geometric and graph reasoning & 2-3 & Tasks that require reasoning about geometric properties, spatial grid navigation, or graph structures, such as shape properties, distances, connectivity, and board-state positioning. &  \begin{minipage}[t]{\linewidth}\begin{itemize}[itemsep=0pt, topsep=0pt, parsep=0pt, partopsep=0pt, leftmargin=15pt]
      \item Please draw a diagonal in a triangle.
      \item Given a directed weighted graph, what is the shortest path from A to B? 
  \end{itemize} \end{minipage}
  \\ \addlinespace[8pt]
   
  & Constraint reasoning & 2-4 & Tasks that require satisfying multiple hard constraints within a complex search space; often equivalent to constraint satisfaction, feasibility checking, finding optimal tactical sequences where most possible paths are invalidated by the rules. &\begin{minipage}[t]{\linewidth}
  \begin{itemize}[itemsep=0pt, topsep=0pt, parsep=0pt, partopsep=0pt, leftmargin=15pt]
      \item Solve the attached sudoku please.
  \end{itemize} \end{minipage}\\ \addlinespace[8pt]
\midrule

\multirow{3}{*}{\textbf{Language and knowledge}}
  & Irrelevant-context robustness & 3-1 &  Tasks where irrelevant or misleading context interferes with correct reasoning. & \begin{minipage}[t]{\linewidth}\begin{itemize}[itemsep=0pt, topsep=0pt, parsep=0pt, partopsep=0pt, leftmargin=15pt]
      \item Imagine you are in a dark room. In a drawer there are 10 red socks, 10 white socks, and 10 green socks. The light comes on. How many socks do you need to take out to be sure you have at least one pair of the same color?
  \end{itemize} \end{minipage}\\ \addlinespace[8pt]

  & Character-level manipulation
  & 3-2 & Represent, process, and manipulate text at the atomic granularity of character as opposed to higher-level tokens.

  &\begin{minipage}[t]{\linewidth} \begin{itemize}[itemsep=0pt, topsep=0pt, parsep=0pt, partopsep=0pt, leftmargin=15pt]
      \item Repeat the number ``12341234'' exactly 31 times without spaces. 
      \item Write a sentence that ends with the letter ``m''.
  \end{itemize} \end{minipage}\\ \addlinespace[8pt]

  & World knowledge
  & 3-3 & Requires factual knowledge (often stable world facts) and avoiding hallucinated details. Or Requires everyday commonsense inferences beyond explicit text (e.g., typical object uses, social/physical plausibility). & 
  \begin{minipage}[t]{\linewidth}\begin{itemize}[itemsep=0pt, topsep=0pt, parsep=0pt, partopsep=0pt, leftmargin=15pt]
      \item  Draw a calendar for February with 29 days. 
      \item What has no holes and holds water? 
  \end{itemize} \end{minipage}
 \\ \addlinespace[8pt]
\bottomrule
\end{tabular}
\end{adjustbox}
\end{table}


\subsection{Task type analsysis}
\label{app:task_type_analyses}
We provide detailed description and examples of task taxonomy in \autoref{tab:dataset_category}.
Beyond the distribution of question format and task taxonomy, we examine the resulting distribution and observe several recurring patterns in the designed challenging tasks. Since the prompts were intentionally created to probe the weaknesses of modern AI agents, these frequently appearing categories may offer useful clues about the capabilities that remain difficult for current models. Below, we provide a brief dataset-level analysis and link the possible reasons to several related studies for easier future study.

\textbf{Object-centric tasks:} We observe a failure for questions related to attribute and pattern recognition  because models tend to extract and report frequently occurring patterns from their training dataset distribution rather than perceiving the actual content as reported in ~\cite{vo2025vision,mitrevski2025inksight,inkslop2026}. In attribute binding or generative counting, they could fail in producing novel recombinations of familiar concepts that rarely occur in natural data. This challenge has been emphasized in prior work ~\cite{wiedemer2023compositional,keysers2019measuring,redhardt2025scaling,binyamin2025make}.  Another example in perceptual counting~\citep{zhang2018learning,acharya2019tallyqa}, models often misidentify, merge or overlook objects in cluttered scenes, especially when instances are small, visually similar, or partially occluded. Spatial reasoning has also been widely recognized as a challenging capability~\cite{spatialab,liu2025spatial} for current models, as it requires understanding and manipulating spatial relations rather than merely recognizing objects. While humans can usually solve these problems through intuitive geometric and physical reasoning, models may fail when the task requires precise relational understanding or multi-step spatial manipulation. In our taxonomy, we therefore include spatial reasoning as a key subcategory of object-centric tasks, covering questions that test whether models can correctly interpret, transform, or reason about spatial configurations.

\textbf{Abstract Reasoning:}
In abstract reasoning, model often fails in problems that require several logical steps and reveals the reasoning gap where the accuracy of the model decreases with the increase of number of inference step. This is justified in \cite{saparov2022language} , where they show that due to the greedy nature of the model, it makes choices locally at each step without a global verification. This leads to accumulation of the error and fails in giving a final correct solution. Moreover, on the problems with constraint reasoning, where strict rules should be followed, or graph ones, such as the maze problems, where a certain path needs to be found, the model suffers from backtracking as mentioned in \cite{valmeekam2023planbench}. In this case, it cannot effectively re-evaluate earlier decision when a path leads to a dead-end or violates a certain constraint and ends up in hallucinated answer.

\textbf{Language and Knowledge tasks:}
For Character-level manipulation, we observe failure in several tasks , such as generating an exact number of characters or repeating a token sequence under strict constraints, especially in settings where external tools are disallowed. We conjecture that such failures may often be attributed to limitations of subword tokenization in large language models, making exact character-level counting and length control  difficult~\cite{zhang2024counting,fu2024large}. Also, models are not robust to irrelevant changes in the input. Prior work \cite{veitch2021counterfactual} has studied this phenomenon as spurious correlations, where model predictions depend on features that are not causally relevant to the task.

\section{Full Grader Prompts}
\label{appendix:full_prompts}
\begin{promptbox}[Grader Prompt (text generation tasks)]
You are an expert grader. Your task is to evaluate the correctness of a submitted answer based on the provided question and ground truth answer defining grading criteria. \\

---

\# [Question] \\
\{question\} \\

---

\# [Submission] \\
\{answer\} \\

---

\# [Ground Truth] \\
\{criterion\} \\

---

Important grading rules: \\
- The submission is correct only if it clearly states a final answer to the question. \\
- A submission that only contains code, tool calls, search queries, reasoning, or instructions for how to derive the answer must be graded 'I' if that answer is not clearly reported at the end, even if those steps would likely produce the correct answer. \\
- If the submission is ambiguous about what the final answer is, grade it 'I'. \\
\\
After assessing the submitted answer, reply with 'GRADE: \$LETTER' (without quotes) where LETTER is either of C or I. Please choose ONE option for the grade: either "C" for correct answers, or "I" for incorrect answers. No intermediate grades are allowed. If the grading criterion is met only partially or the ground truth is ambiguous, use your best judgement to assign the most appropriate grade.\\
\\
Start by briefly analyzing the submission and compare it against the ground truth. Be as concise as possible in your analysis. The ground truth will provide you with the necessary information to easily determine if the submission is correct or incorrect. Only if needed, execute code snippets to verify the correctness of the submission. \\
Write a minimal explanation of your judgement and end with your answer formatted as 'GRADE: \$LETTER' (without quotes) where LETTER is either of C or I.
\end{promptbox}

\begin{promptbox}[Grader Prompt (image generation tasks)]
You are an expert grader evaluating whether a generated image correctly satisfies the given task.

---

\# [Task/Prompt] \\
\{question\} \\
\{optional-input-image-section\} \\

---

\# [Generated Image (Candidate Answer to Grade)] \\
The image below is the generated output that needs to be evaluated. (The generated image is attached after this text) \\

---

\# [Grading Criteria] \\
\{criterion\} \\

---

\# [Instructions] \\
Evaluate whether the generated image correctly satisfies the task requirements based on the grading criteria above.

Consider the following aspects when grading: \\
1. **Content Accuracy**: Does the image contain the required elements/objects/scenes? \\
2. **Instruction Following**: Does the image follow the specific instructions in the prompt? \\
3. **Quality**: Is the generated image of reasonable quality (not corrupted, incomplete, or nonsensical)? \\
4. **Relevance**: Is the generated image relevant to the task? \\
\\
If needed, you can execute Python code snippets to help verify aspects of the generated image (e.g., counting objects, checking colors, analyzing dimensions, etc.).

After your analysis, provide your final grade. \\
Reply with 'GRADE: \$LETTER' (without quotes) where LETTER is either: \\
- "C" for CORRECT: The generated image satisfies the task requirements \\
- "I" for INCORRECT: The generated image does NOT satisfy the task requirements \\

Start with a brief analysis of the generated image compared to the requirements, then provide your final grade.
\end{promptbox}

\section{Grader Validation}
\autoref{tab:pipeline_validation} and \cref{tab:pipeline_validation_bias} collect the detailed metrics computed to verify the alignment between the automatic evaluation pipeline and human judgment.

\begin{table}[t]
\caption{Pipeline Validation Results: agreement scores between human-annotated correctness scores and values obtained with the automatic evaluation pipeline.}
\label{tab:pipeline_validation}
\centering
\small
\begin{tabular}{lccccccccc}
\toprule
Output type & $n$ & \multicolumn{2}{c}{Grader} & \multicolumn{2}{c}{Human} & \textbf{Acc.} & \textbf{Prec.} & \textbf{Rec.} & \textbf{F1} \\
\cmidrule(lr){3-4}\cmidrule(lr){5-6}
 & & Correct & Incorrect & Correct & Incorrect & & & &  \\
\midrule
Text & 88 & 48 & 40 & 47 & 41 & 0.966 & 0.958 & 0.979 & 0.968 \\
Image & 66 & 33 & 33 & 31 & 35 & 0.909 & 0.879 & 0.935 & 0.906 \\
\bottomrule
\end{tabular}
\end{table}

\begin{table}[t]
\centering
\caption{
Disaggregated grader validation by solver model family. We report agreement between the automatic grader and human labels, together with false positive rate (FPR) and recall. FPR measures the rate at which the grader incorrectly marks a human-incorrect response as correct, and is therefore the most relevant metric for detecting potential overestimation of a model family's performance.
}
\label{tab:pipeline_validation_bias}
\begin{adjustbox}{max width=\textwidth}
\begin{tabular}{llrrrrrr}
\toprule
Output type & Solver subset & $n$ & Human correct & Grader correct & \textbf{Accuracy} & \textbf{FPR} & \textbf{Recall} \\
\midrule
Text & Google & 21 & 13 & 12 & 0.952 & 0.000 & 0.923 \\
 & Others & 67 & 34 & 36 & 0.970 & 0.061 & 1.000 \\
\midrule
Image & Google & 33 & 18 & 17 & 0.909 & 0.067 & 0.889 \\
 & Others & 33 & 13 & 16 & 0.909 & 0.150 & 1.000 \\
\bottomrule
\end{tabular}
\end{adjustbox}
\end{table}

\section{Models Info List}
\label{appendix:models_info}
\autoref{tab:models-serving-hardware} collects all the specifics of the evaluated models.

\begin{table}[]
\caption{Model metadata, pricing, datatype, and serving hardware on the serving platform. Prices are reported in USD per 1M tokens. Size is reported in billions of parameters when available. Serving GPU counts and data-types are taken from the serving platform default model configuration.
\textbf{VLMs} can process both images and textual inputs but only produce text, while \textbf{LLM} only operate in text. \textbf{Image generation} models can instead processed both modalities as input, but their output are only images.}
\small
\label{tab:models-serving-hardware}
\begin{tabular}{lrrrrll}
\toprule
\textbf{Model} &
\textbf{Size (B)} &
\textbf{Release} &
\textbf{In. (\$)} &
\textbf{Out. (\$)} &
\textbf{dtype} &
\textbf{GPUs} \\
\midrule
\multicolumn{7}{l}{\textbf{Language models}} \\
\midrule
GLM-4.7 & 358 & 2025-12 & 0.5606 & 1.6817 & fp8 & 4 x H200 \\
GLM-5 & 754 & 2026-02 & 1.0758 & 3.2274 & fp8 & 8 x H200 \\
GLM-5.1 & 754 & 2026-04 & 0.7716 & 2.3149 & fp8 & 8 x H200 \\
GLM-5.2 & 754 & 2026-06 & 0.69 & 2.07 & fp8 & 8 x H200 \\
DeepSeek-V4-Flash & 158 & 2026-04 & 0.2666 & 0.7998 & fp8 & 4 x H100 \\
DeepSeek-V4-Pro & 862 & 2026-04 & 0.9324 & 2.7971 & fp8 & 8 x H200 \\
GPT-oss-20b & 21 & 2025-08 & 0.0204 & 0.0611 & bf16 & 1 x A100 \\
GPT-oss-120b & 117 & 2025-08 & 0.0486 & 0.1457 & bf16 & 1 x A100 \\
Qwen3-Next-80B-A3B-Thinking & 80 & 2025-09 & 0.1365 & 0.4097 & bf16 & 4 x A100 \\
\midrule
\multicolumn{7}{l}{\textbf{Vision-language models}} \\
\midrule
Kimi-K2.5 & 1000 & 2026-02 & 0.6856 & 2.0567 & int4 & 8 x H200 \\
Kimi-K2.6 & 1000 & 2026-04 & 0.6121 & 1.8362 & int4 & 8 x H200 \\
Gemma-4-E2B-it & 5 & 2026-04 & 0.0164 & 0.0493 & bf16 & 1 x A100 \\
Gemma-4-E4B-it & 8 & 2026-04 & 0.0249 & 0.0746 & bf16 & 1 x A100 \\
Gemma-4-26B-A4B-it & 26 & 2026-04 & 0.0422 & 0.1266 & bf16 & 1 x A100 \\
Qwen3-VL-30B-A3B-Thinking & 30 & 2025-09 & 0.0633 & 0.1902 & bf16 & 1 x A100 \\
Qwen3-VL-235B-A22B-Thinking & 235 & 2025-09 & 0.3095 & 0.9285 & fp8 & 4 x A100 \\
Qwen3.5-35B-A3B & 35 & 2026-03 & 0.0483 & 0.1448 & fp8 & 1 x A100 \\
Qwen3.5-122B-A10B & 122 & 2026-03 & 0.1491 & 0.4474 & fp8 & 2 x A100 \\
Qwen3.5-397B-A17B & 397 & 2026-03 & 0.5047 & 1.5142 & fp8 & 8 x A100 \\
Gemini-2.5-flash & -- & 2025-06 & 0.3000 & 2.5000 & -- & -- \\
Gemini-2.5-pro & -- & 2025-06 & 1.2500 & 10.0000 & -- & -- \\
Gemini-3-flash-preview & -- & 2025-12 & 0.5000 & 3.0000 & -- & -- \\
Gemini-3.1-flash-lite-preview & -- & 2026-03 & 0.2500 & 1.5000 & -- & -- \\
Gemini-3.1-pro-preview & -- & 2026-02 & 1.0000 & 6.0000 & -- & -- \\
GPT-5-mini & -- & 2025-08 & 0.2500 & 2.0000 & -- & -- \\
GPT-5 & -- & 2025-08 & 1.2500 & 10.0000 & -- & -- \\
GPT-5.2 & -- & 2025-12 & 1.7500 & 14.0000 & -- & -- \\
GPT-5.4-nano & -- & 2026-03 & 0.2000 & 1.2500 & -- & -- \\
GPT-5.4-mini & -- & 2026-03 & 0.7500 & 4.5000 & -- & -- \\
GPT-5.4 & -- & 2026-03 & 2.5000 & 15.0000 & -- & -- \\
GPT-5.5 & -- & 2026-04 & 5.0000 & 30.0000 & -- & -- \\
\midrule
\multicolumn{7}{l}{\textbf{Generative image models}} \\
\midrule
Gemini-2.5-flash-image & -- & 2025-08 & 0.3000 & 30.0000 & -- & -- \\
Gemini-3-pro-image-preview & -- & 2025-11 & 2.0000 & 120.0000 & -- & -- \\
Gemini-3.1-flash-image-preview & -- & 2026-03 & 0.5000 & 60.0000 & -- & -- \\
GPT-image-1-mini & -- & 2025-10 & 2.5000 & 8.0000 & -- & -- \\
GPT-image-1.5 & -- & 2025-12 & 8.0000 & 32.0000 & -- & -- \\
GPT-image-2 & -- & 2026-04 & 8.0000 & 30.0000 & -- & -- \\
\bottomrule
\end{tabular}
\normalsize

\end{table}

\section{Full Models Performance}
\label{appendix:full_accuracy}
\begin{figure}
    \centering
    \includegraphics[width=\linewidth]{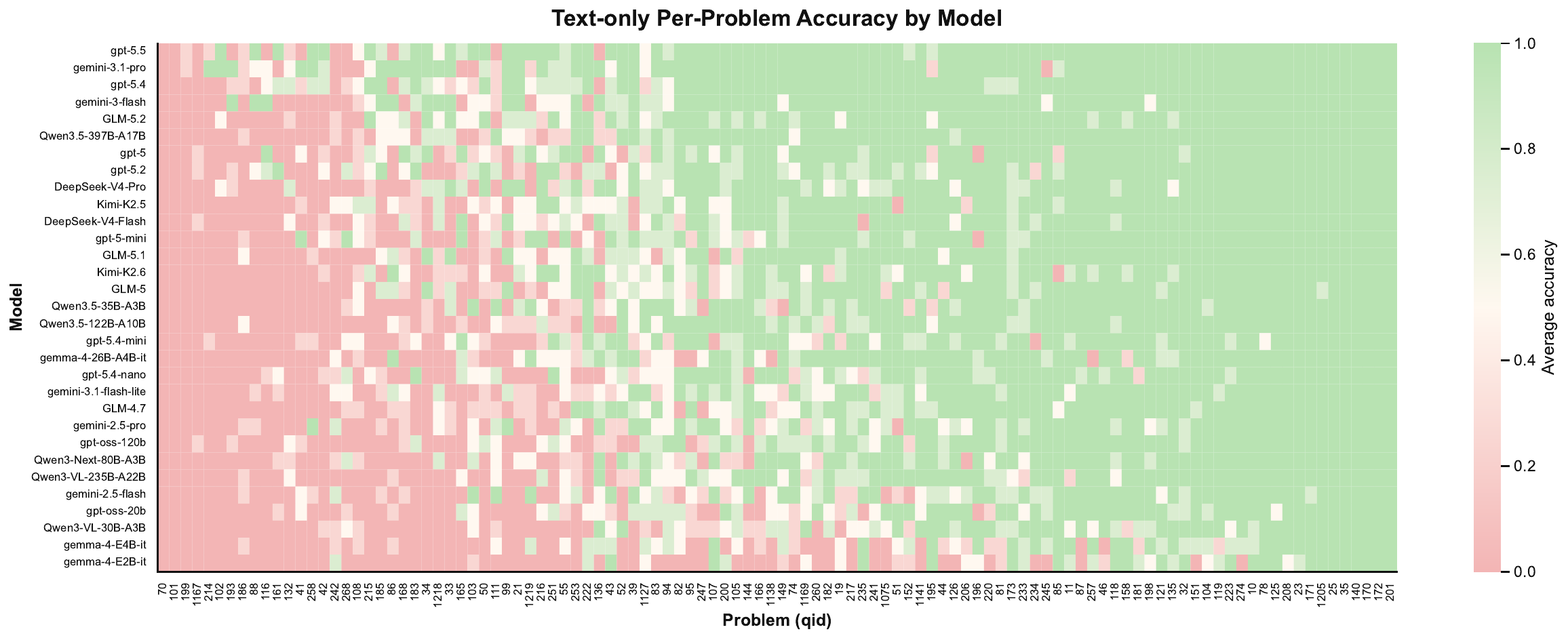}
    \caption{Average accuracy for each model on each task, on \textbf{text-only} problems.}
    \label{fig:qid_acc_text}
\end{figure}

\begin{figure}
    \centering
    \includegraphics[width=0.7\linewidth]{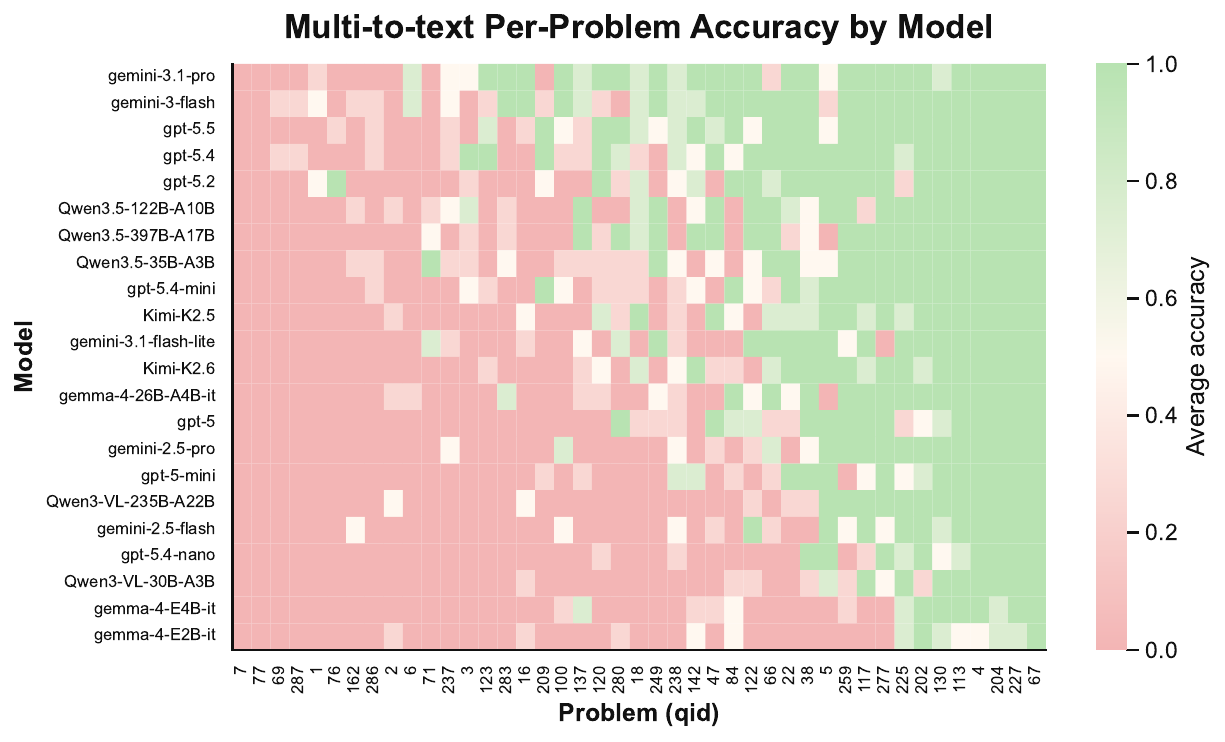}
    \caption{Average accuracy for each model on each task, on \textbf{multi-to-text} problems.}
    \label{fig:qid_acc_multi}
\end{figure}

\begin{figure}
    \centering
    \includegraphics[width=\linewidth]{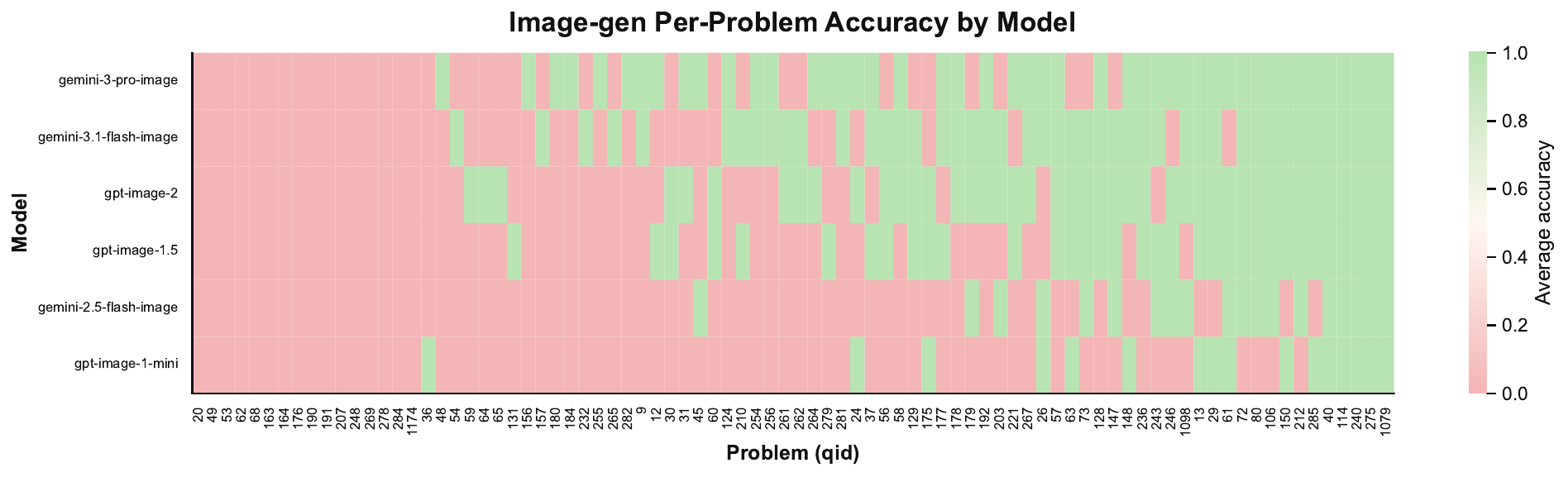}
    \caption{Average accuracy for each model on each task, on \textbf{image-generation} problems.}
    \label{fig:qid_acc_imggen}
\end{figure}

Hereafter are reported the scores of all models on each single question, allowing to identify shared failures among models. Three heatmaps collect the models mean@k for each problem. Problems are sorted based on the average accuracy across models. Results are reported in \cref{fig:qid_acc_text} for text-only, \cref{fig:qid_acc_multi} for multi-to-text, and \cref{fig:qid_acc_imggen} for image generation.

\section{Additional analysis}
Analyses relying on the proposed taxonomy are presented in \autoref{tab:analyses_category}.
\begin{table}[htbp!]
\centering
\caption{Detailed accuracy performance of each model on individual subtask categories. The mapping between task IDs and task names can be found in \autoref{tab:dataset_category}.}
\label{tab:analyses_category}
\setlength{\tabcolsep}{3pt}
\begin{adjustbox}{max width=\textwidth}
\begin{tabular}{llllllllllllll}
\toprule
\textbf{Model name} & \textbf{Model type} & \textbf{1-1} & \textbf{1-2} & \textbf{1-3} & \textbf{1-4} & \textbf{1-5} & \textbf{2-1} & \textbf{2-2} & \textbf{2-3} & \textbf{2-4} & \textbf{3-1} & \textbf{3-2} & \textbf{3-3}  \\
\midrule
\bf Total count & ALL & 6 & 53 & 18 & 21 & 30 & 18 & 22 & 18 & 9 & 8 & 32 & 19  \\
\midrule
\bf Total count & GEN & 0 & 37 & 18 & 0 & 30 & 1 & 0 & 7 & 2 & 1 & 0 & 6  \\
\hdashline
gemini-2.5-flash-image & GEN & - & 29.73 & 38.89 & - & 40.00 & 0.00 & - & 0.00 & 0.00 & 100.00 & - & 50.00   \\
gemini-3-pro-image-preview & GEN & - & 54.05 & 61.11 & - & 73.33 & 0.00 & - & 42.86 & 50.00 & 100.00 & - & 66.67 \\
gemini-3.1-flash-image-preview & GEN & - & 64.86 & 55.56 & - & 60.00 & 0.00 & - & 57.14 & 0.00 & 100.00 & - & 83.33   \\
gpt-image-1-mini & GEN & - & 21.62 & 33.33 & - & 46.67 & 0.00 & - & 0.00 & 0.00 & 100.00 & - & 0.00   \\
gpt-image-1.5 & GEN & - & 51.35 & 50.00 & - & 63.33 & 0.00 & - & 14.29 & 0.00 & 100.00 & - & 33.33  \\
gpt-image-2 & GEN & - & 54.05 & 66.67 & - & 66.67 & 0.00 & - & 14.29 & 0.00 & 100.00 & - & 83.33   \\
\midrule
\bf Total count & LM & 1 & 10 & 0 & 1 & 0 & 17 & 18 & 7 & 6 & 7 & 31 & 11 \\
\hdashline
zai-org/GLM-4.7 & LM & 0.00 & 62.50 & - & 0.00 & - & 67.65 & 68.06 & 71.43 & 37.50 & 64.29 & 54.84 & 68.18   \\
zai-org/GLM-5 & LM & 75.00 & 60.00 & - & 0.00 & - & 57.35 & 66.67 & 78.57 & 41.67 & 85.71 & 57.26 & 84.09  \\
zai-org/GLM-5.1 & LM & 50.00 & 62.50 & - & 0.00 & - & 54.41 & 63.89 & 85.71 & 45.83 & 85.71 & 67.74 & 86.36   \\
zai-org/GLM-5.2 & LM & 50.00 & 72.50 & - & 0.00 & - & 75.00 & 73.61 & 85.71 & 41.67 & 78.57 & 73.39 & 84.09 \\
deepseek-ai/DeepSeek-V4-Flash & LM & 0.00  & 62.50 & -- & 0.00 & -- & 73.53 & 68.06 & 82.14 & 45.83 & 57.14 & 72.58 & 63.64    \\
deepseek-ai/DeepSeek-V4-Pro & LM & 0.00 & 65.00 & -- & 0.00 & -- & 64.71 & 73.61 & 82.14 & 33.33 & 71.43 & 75.00 & 84.09   \\
openai/gpt-oss-20b & LM & 0.00 & 55.00 & -- & 0.00 & -- & 51.47 & 59.72 & 71.43 & 33.33 & 53.57 & 52.42 & 40.91  \\
openai/gpt-oss-120b & LM & 0.00 & 55.00 & -- & 0.00 & -- & 66.18 & 66.67 & 85.71 & 50.00 & 57.14 & 59.68 & 43.18   \\
Qwen/Qwen3-Next-80B-A3B-Thinking & LM & 0.00 & 62.50 & - & 0.00 & -  & 66.18 & 66.67 & 82.14 & 33.33 & 50.00 & 56.45 & 54.55   \\
\midrule
\bf Total count & VLM & 6 & 16 & 0 & 21 & 0 & 17 & 22 & 11 & 7 & 7 & 32 & 13   \\
\hdashline
moonshotai/Kimi-K2.5 & VLM & 20.83 & 54.69 & -- & 33.33 & -- & 76.47 & 82.95 & 70.45 & 53.57 & 60.71 & 64.06 & 57.69 \\
moonshotai/Kimi-K2.6 & VLM & 12.50 & 51.56 & -- & 34.52 & -- & 63.24 & 75.00 & 70.45 & 46.43 & 50.00 & 69.53 & 61.54    \\
google/gemma-4-E2B-it & VLM & 4.17 & 39.06 & -- & 9.52 & -- & 27.94 & 46.59 & 27.27 & 21.43 & 50.00 & 25.00 & 25.00    \\
google/gemma-4-E4B-it & VLM & 0.00 & 31.25 & -- & 16.67 & -- & 52.94 & 57.95 & 29.55 & 25.00 & 57.14 & 34.38 & 32.69  \\
google/gemma-4-26B-A4B-it & VLM & 4.17 & 56.25 & -- & 34.52 & -- & 69.12 & 75.00 & 61.36 & 39.29 & 42.86 & 66.41 & 50.00  \\
Qwen/Qwen3-VL-30B-A3B-Thinking & VLM & 0.00 & 35.94 & -- & 17.86 & -- & 55.88 & 67.05 & 54.55 & 14.29 & 39.29 & 51.56 & 32.69  \\
Qwen/Qwen3-VL-235B-A22B-Thinking & VLM & 8.33 & 43.75 & -- & 25.00 & -- & 64.71 & 68.18 & 54.55 & 28.57 & 53.57 & 60.16 & 53.85  \\
Qwen/Qwen3.5-35B-A3B & VLM & -- & 51.56 & -- & 47.62 & -- & 75.00 & 79.55 & 72.73 & 32.14 & 75.00 & 55.47 & 53.85  \\
Qwen/Qwen3.5-122B-A10B & VLM & 29.17 & 51.56 & -- & 53.57 & -- & 64.71 & 69.32 & 59.09 & 39.29 & 75.00 & 65.62 & 53.85   \\
Qwen/Qwen3.5-397B-A17B & VLM & 25.00 & 60.94 & -- & 55.95 & -- & 72.06 & 71.59 & 65.91 & 39.29 & 82.14 & 67.97 & 69.23  \\
gemini-2.5-flash & VLM & 4.17 & 45.31 & -- & 27.38 & -- & 70.59 & 72.73 & 63.64 & 21.43 & 60.71 & 41.41 & 48.08\\
gemini-2.5-pro & VLM & 4.17 & 54.69 & -- & 29.76 & -- & 55.88 & 68.18 & 68.18 & 32.14 & 71.43 & 53.91 & 69.23\\
gemini-3-flash-preview & VLM & 41.67 & 75.00 & -- & 59.52 & -- & 76.47 & 72.73 & 88.64 & 53.57 & 71.43 & 77.34 & 82.69  \\
gemini-3.1-flash-lite-preview & VLM & 4.17 & 60.94 & - & 40.48 & - & 57.35 & 68.18 & 61.36 & 28.57 & 60.71 & 64.06 & 63.46   \\
gemini-3.1-pro-preview & VLM & 41.67 & 87.50 & - & 57.14 & - & 92.65 & 75.00 & 88.64 & 64.29 & 96.43 & 84.38 & 76.92   \\
gpt-5-mini & VLM & 0.00 & 53.12 & - & 26.19 & - & 77.94 & 79.55 & 75.00 & 32.14 & 64.29 & 68.75 & 42.31   \\
gpt-5 & VLM & 12.50 & 59.38 & - & 38.10 & - & 77.94 & 71.59 & 68.18 & 57.14 & 60.71 & 69.53 & 55.77   \\
gpt-5.2 & VLM & 41.67 & 62.50 & - & 39.29 & - & 70.59 & 76.14 & 75.00 & 71.43 & 64.29 & 71.09 & 59.62 \\
gpt-5.4-nano & VLM & 4.17 & 48.44 & - & 16.67 & - & 72.06 & 72.73 & 54.55 & 46.43 & 53.57 & 64.06 & 34.62  \\
gpt-5.4-mini & VLM & 12.50 & 50.00 & - & 36.90 & - & 60.29 & 75.00 & 75.00 & 35.71 & 75.00 & 66.41 & 53.85 \\
gpt-5.4 & VLM & 41.67 & 70.31 & - & 52.38 & - & 85.29 & 77.27 & 81.82 & 64.29 & 89.29 & 81.25 & 63.46   \\
gpt-5.5 & VLM & 25.00 & 73.44 & - & 51.19 & - & 89.71 & 88.64 & 84.09 & 60.71 & 100.00 & 82.03 & 80.77 \\
\bottomrule

\end{tabular} 
\end{adjustbox}
\end{table}
\begin{table}[htbp!]
\centering
\caption{Detailed accuracy performance of each model on individual task categories.}
\label{tab:analyses_category-2}
\setlength{\tabcolsep}{3pt}
\begin{adjustbox}{max width=\textwidth}
\begin{tabular}{lllll}
\toprule
\textbf{Model} & \textbf{Model type} & \textbf{Object-centric} & \textbf{Abstract Reasoning} & \textbf{Language and Knowledge} \\
\midrule
\bf ALL & - &  128 & 67 & 59  \\
\midrule
\bf Total count & GEN &  85 & 10 & 7  \\
\hdashline
gemini-2.5-flash-image & GEN &  35.29 & 0.00 & 57.14 \\
gemini-3-pro-image-preview & GEN &  62.35 & 40.00 & 71.43\\
gemini-3.1-flash-image-preview & GEN & 61.18 & 40.00 & 85.71  \\
gpt-image-1-mini & GEN &  32.94 & 0.00 & 14.29  \\
gpt-image-1.5 & GEN &  55.29 & 10.00 & 42.86  \\
gpt-image-2 & GEN & 61.18 & 10.00 & 85.71  \\
\midrule
\bf Total count & LM & 12 & 48 & 49 \\
\hdashline
zai-org/GLM-4.7 & LM & 52.08 & 64.58 & 59.18  \\
zai-org/GLM-5 & LM &  56.25 & 61.98 & 67.35 \\
zai-org/GLM-5.1 & LM & 56.25 & 61.46 & 74.49 \\
zai-org/GLM-5.2 & LM & 64.58 & 71.88 & 76.53 \\
deepseek-ai/DeepSeek-V4-Flash & LM & 52.08 & 69.27 & 68.37  \\
deepseek-ai/DeepSeek-V4-Pro & LM &  54.17 & 66.67 & 76.53  \\
openai/gpt-oss-20b & LM & 45.83 & 55.21 & 50.00  \\
openai/gpt-oss-120b & LM & 45.83 & 67.19 & 55.61  \\
Qwen/Qwen3-Next-80B-A3B-Thinking & LM & 52.08 & 64.58 & 55.10  \\
\midrule
\bf Total count &VLM & 43 & 57 & 52 \\
\hdashline
moonshotai/Kimi-K2.5 & VLM  & 39.53 & 75.00 & 62.02  \\
moonshotai/Kimi-K2.6 & VLM & 37.79 & 67.11 & 64.90  \\
google/gemma-4-E2B-it & VLM & 19.77 & 34.21 & 28.37  \\
google/gemma-4-E4B-it & VLM &  19.77 & 46.93 & 37.02  \\
google/gemma-4-26B-A4B-it & VLM & 38.37 & 66.23 & 59.13  \\
Qwen/Qwen3-VL-30B-A3B-Thinking & VLM &22.09 & 54.82 & 45.19  \\
Qwen/Qwen3-VL-235B-A22B-Thinking & VLM & 29.65 & 59.65 & 57.69  \\
Qwen/Qwen3.5-35B-A3B & VLM & 43.60 & 71.05 & 57.69  \\
Qwen/Qwen3.5-122B-A10B & VLM & 49.42 & 62.28 & 63.94  \\
Qwen/Qwen3.5-397B-A17B & VLM & 53.49 & 66.67 & 70.19 \\
gemini-2.5-flash & VLM &30.81 & 64.04 & 45.67 \\
gemini-2.5-pro & VLM &  35.47 & 60.09 & 60.10 \\
gemini-3-flash-preview & VLM & 62.79 & 74.56 & 77.88  \\
gemini-3.1-flash-lite-preview & VLM & 43.02 & 58.77 & 63.46  \\
gemini-3.1-pro-preview & VLM &  66.28 & 81.58 & 84.13 \\
gpt-5-mini & VLM &32.56 & 72.37 & 61.54 \\
gpt-5 & VLM & 42.44 & 71.05 & 64.90  \\
gpt-5.2 & VLM & 48.26 & 73.68 & 67.31\\
gpt-5.4-nano & VLM & 26.74 & 65.79 & 55.29  \\
gpt-5.4-mini & VLM  & 38.37 & 65.79 & 64.42 \\
gpt-5.4 & VLM & 57.56 & 78.95 & 77.88  \\
gpt-5.5 & VLM  & 55.81 & 84.65 & 84.13 \\
\bottomrule

\end{tabular} 
\end{adjustbox}
\end{table}

\section{Broader Impact}
\label{appendix:broader_impact}
Firstly, we provide a diagnostic benchmark to detect persistent gaps in modern models' language understanding. By selecting tasks that can be easily solved by humans but remain difficult for modern AI, \bench can help to better understand cases of failure which are otherwise overlooked due to aggregate success in other benchmarking tasks. In certain use-cases, where minor mistakes in number or spatial reasoning, attribute binding, or instruction following could lead to misinterpretations or even unsafe actions on the part of an AI system, \bench may become relevant for evaluating the applicability of the models in question, including applications to education, assistive technologies, document parsing, and human-AI collaboration workflows.

The \bench may prove useful as a tool for more insightful model comparisons, showing how different approaches behave within individual categories. At the same time, it would be unwise to consider the proposed benchmark as a means for thorough reliability assessment. Being relatively small in terms of scale, biased towards particular types of problems and limited in question formulation and answer distribution, the dataset will inevitably show some signs of bias when it comes to the overall reliability of various model architectures. Another problem associated with public benchmarks is that in the future, models can be specifically trained to overfit to the given data. For all that, the proposed benchmark can only serve as a stress-test and not a thorough safety evaluation of model robustness.

\section{License information}
We distribute \bench, containing the benchmark prompts, reference solutions, task and sub-task annotations, failure-mode annotations, and associated benchmark metadata for research purposes. This dataset is available on HuggingFace under CC-BY-4.0 license. This data set contains questions suggested by students in an AI course at a graduate level and refined, filtered, and annotated by the authors. The provided annotations consist of the structured reference solution, correctness conditions, and typical failure modes. Some samples include or refer to images and external links to interact with shared models; the respective external service and third-party contents accessible via such links are governed by their own license terms and are not being re-licensed by the benchmark.

The evaluation code and scripts are provided separately from the data set. The evaluation pipeline uses Inspect AI, available under the MIT License. All the third-party software packages used here retain their original license agreements. Outputs from models, evaluation logs, and generated images are not included with the distribution of the benchmark.

\clearpage
\end{document}